\documentclass{article}

\usepackage{xcolor}
\usepackage[final]{corl_2026}
\usepackage[OT1]{fontenc}

\usepackage{graphicx}
\usepackage{booktabs}
\usepackage{multirow}
\usepackage{subcaption}
\usepackage{amsmath, amssymb}

\usepackage[capitalize,noabbrev]{cleveref}
\usepackage{microtype}
\usepackage{wrapfig}
\usepackage{hyperref}

\usepackage{longtable}
\usepackage[most]{tcolorbox}
\newtcolorbox{promptbox}[1][]{%
  enhanced, breakable,
  colback=gray!5, colframe=gray!50,
  boxrule=0.5pt, arc=2pt,
  left=6pt, right=6pt, top=3pt, bottom=3pt,
  coltitle=black, colbacktitle=gray!20,
  fonttitle=\bfseries\small\sffamily,
  title={#1},
  fontupper=\small\ttfamily}

\newcommand{\mainsec}[1]{Section~#1 of the main paper}

\newcommand{\cmark}{\ensuremath{\checkmark}}
\newcommand{\xmark}{\ensuremath{\times}}

\title{MobiAgent: Dual-Loop Recursive Policy Self-\\Improvement for Long-Horizon Mobile Manipulation}

\author{
Chenzhi Liu\textsuperscript{1,*}\quad
Yue Zhang\textsuperscript{1,*}\quad
Jiehong Lin\textsuperscript{1,*\ensuremath{\dagger}}\\
\bfseries Jianan Wang\textsuperscript{2}\quad
Bo Wang\textsuperscript{1}\quad
Zhongrui Wang\textsuperscript{3,\ensuremath{\ddagger}}\quad
Xiaojuan Qi\textsuperscript{1,\ensuremath{\ddagger}}\\[4pt]
\textsuperscript{1}The University of Hong Kong\quad
\textsuperscript{2}Astribot\\[2pt]
\textsuperscript{3}Southern University of Science and Technology\\[4pt]
{\normalfont\small
\textsuperscript{*}Equal contribution\quad
\textsuperscript{\ensuremath{\dagger}}Project Leader\quad
\textsuperscript{\ensuremath{\ddagger}}Corresponding authors
}\\[4pt]
{\normalfont\small
Project Page:
\href{https://dual-loop-agent.github.io}
{\textnormal{https://kaiknower.github.io/mobiagent}}
}\\[2pt]
{\normalfont\small
GitHub:
\href{https://github.com/kaiknower/MobiAgent}
{\textnormal{https://github.com/kaiknower/MobiAgent}}
}
}

\hypersetup{
  pdftitle={From Execution to Evolution: A Dual-Loop Agentic System for Long-Horizon Mobile Manipulation},
  pdfauthor={Chenzhi Liu, Yue Zhang, Jiehong Lin, Jianan Wang, Bo Wang, Zhongrui Wang, Xiaojuan Qi}
}

\begin{document}
\addtocontents{toc}{\protect\setcounter{tocdepth}{-1}}
\maketitle

\begin{abstract}
Long-horizon mobile manipulation presents significant challenges due to compounding execution errors and capacity interference between locomotion and arm control. While recent Vision-Language-Action models excel at short-horizon tasks, they lack the hierarchical reasoning required for multi-stage objectives. Furthermore, existing hierarchical agents suffer from rigid sub-task mapping, inflexible replanning, and a lack of continuous learning. To address these limitations, we introduce \textbf{MobiAgent}, a dual-loop agentic framework that bridges robust deployment execution and recursive policy self-improvement. During deployment, the \textit{Inner Loop} decouples high-level reasoning from low-level control through highly composable \textit{atomic skills}. It employs Vision-Language models for receding-horizon planning and visual reflection, dynamically composing skills to ensure robust error recovery. These skills are executed by specialized flow-matching experts that share a unified VLM backbone, maximizing reusability while mitigating capacity interference. 
Concurrently, the \textit{Outer Loop} drives automated lifelong learning by autonomously segmenting and verifying deployment rollouts, clustering them to discover atomic skills, and continuously fine-tuning the skill library without human annotations.
Evaluations on RoboCasa, BEHAVIOR-1K, and real-world tasks demonstrate the effectiveness of MobiAgent. It outperforms $\pi_{0.5}$-TA by 22.5 percentage points on BEHAVIOR-1K and enables robust recovery from execution failures. Through autonomous data recycling, success improves from 7.50\% to 27.50\% on RoboCasa and from 32.5\% to 57.5\% on Astribot S1.
\end{abstract}


\section{Introduction}
\label{sec:introduction}

\begin{figure}[t]
 \centering
 \includegraphics[width=1\linewidth]{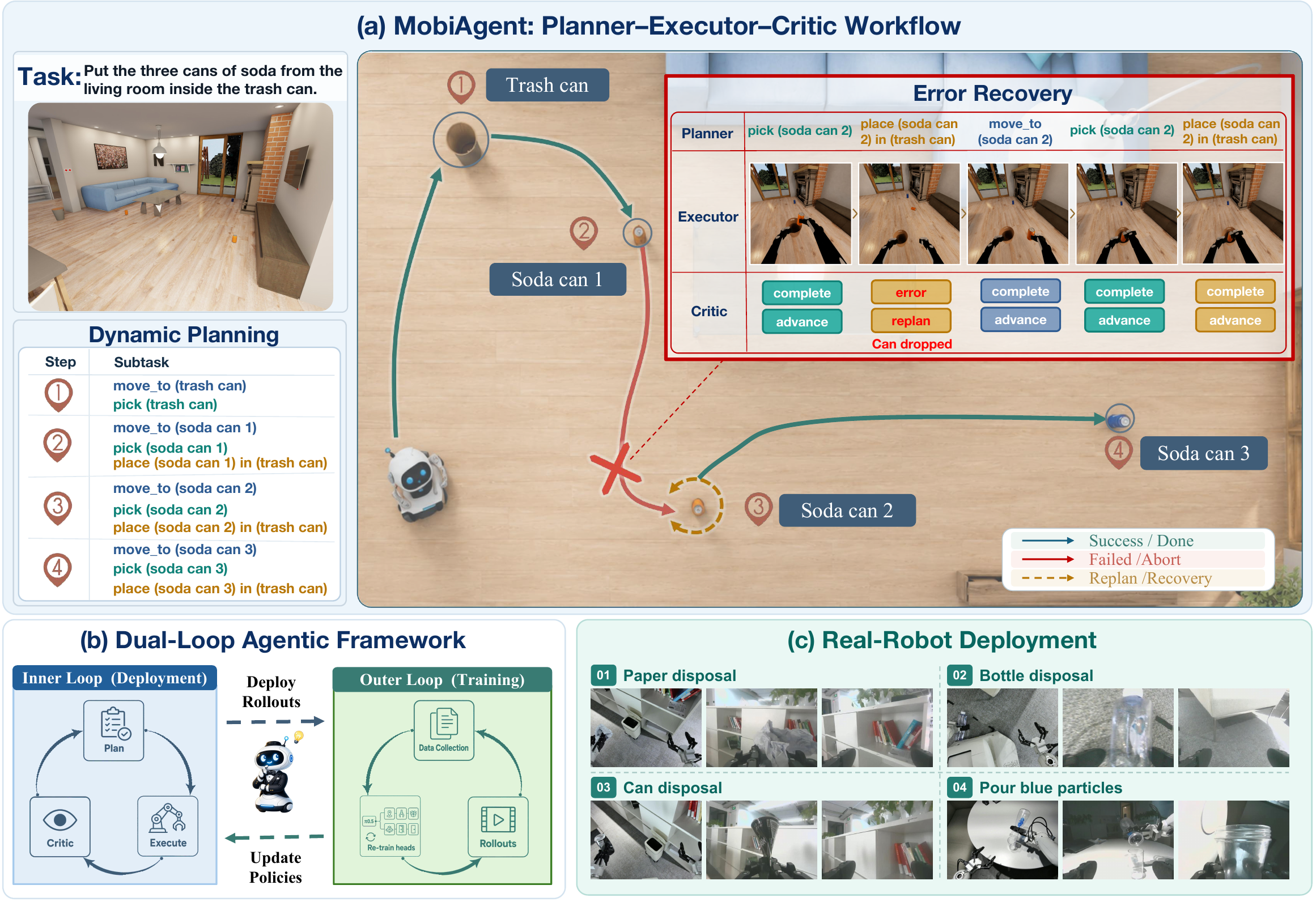}
 \caption{\textbf{Overview of MobiAgent.} 
 MobiAgent solves a long-horizon \texttt{picking\_up\_trash} task by interleaving dynamic planning, skill-conditioned execution, and critic-guided recovery. The rollout trace illustrates a failed placement where the can drops outside the trash. In response, MobiAgent triggers replanning, guiding the robot to approach the displaced can, re-grasp it, and complete the task. This behavior is driven by our dual-loop framework (bottom left), which couples an online \textit{Inner Loop} executing atomic skills with an offline \textit{Outer Loop} that autonomously curates data for continuous policy evolution. The bottom right shows real-robot deployment on paper, bottle, and can disposal, as well as pouring blue particles into a cup.
 }
 \label{fig:teaser}
\end{figure}

Long-horizon mobile manipulation requires executing complex, multi-stage tasks to achieve global goals, necessitating the seamless interleaving of distinct behavioral modes including global navigation and precise prehension. This domain poses significant challenges due to compounding execution errors and the capacity interference inherent in jointly modeling disparate sensorimotor mappings for locomotion and arm control.

Recent Vision-Language-Action (VLA) models, notably OpenVLA~\citep{kim2024openvla}, \(\pi_{0}\)~\citep{physicalintelligence2024pi0}, and \(\pi_{0.5}\)~\citep{physicalintelligence2025pi0_5}, have demonstrated exceptional proficiency in tabletop environments by directly mapping multimodal observations to continuous actions. Building on this foundation, systems like Mobi-\(\pi\)~\citep{yang2025mobipi} and N2M~\citep{chai2025n2m} extend VLAs to mobile platforms, primarily optimizing the navigation-manipulation hand-off. Nevertheless, these monolithic architectures function predominantly as reactive policies tailored for short-horizon objectives. In long-horizon scenarios, they lack the hierarchical reasoning necessary for semantic task decomposition and struggle to maintain the long-term context required for closed-loop replanning, fundamentally limiting their autonomy and reliability.

To transcend short-horizon reactive policies, recent research explores hierarchical agentic frameworks that layer Large Language Models (LLMs) or Vision-Language Models (VLMs) over low-level controllers~\citep{yao2023react, liang2023codeaspolicies, huang2023voxposer, li2024manipllm, wu2024momallm, shi2025hirobot, zhao2025vla2, roboclaw2026}, typically adopting a \textit{plan-execute-verify} paradigm. Beyond their confinement to stationary manipulation, these architectures exhibit three critical bottlenecks: 
\textbf{1) Constrained Planning via Sub-task Mapping:} They typically assign an individual, isolated policy to each decomposed sub-task. High-level planning is then reduced to discrete selection from a predefined list rather than open-ended generation, inherently limiting scalability and increasing system overhead as the required sub-task repertoire expands. 
\textbf{2) Inflexible Verification and Replanning:} Stemming directly from this rigid policy mapping, dynamically replanning after a verification failure becomes highly cumbersome. This inflexibility is particularly fatal in mobile manipulation, where reflection and error recovery demand complex, continuous spatial reasoning (e.g., dynamic navigation, distance and viewpoint adjustments) prior to interaction. 
\textbf{3) Absence of Data Curation and Policy Self-Improvement:} The vast majority of these frameworks operate open-loop regarding lifelong learning. While rare exceptions like RoboClaw~\citep{roboclaw2026} introduce automated data collection, the field still lacks a comprehensive pipeline for unconstrained data curation and recursive self-improvement.

To bridge the gap between robust deployment execution and recursive policy self-improvement, we propose \textbf{MobiAgent}, a dual-loop agentic system tailored for long-horizon mobile manipulation, as shown in Fig. \ref{fig:teaser}. MobiAgent decouples high-level reasoning from low-level control based on atomic skills, partitioning the system into a deployment-time Inner Loop and an offline-training Outer Loop.

To overcome the bottlenecks of constrained planning and inflexible replanning, the Inner Loop shifts the execution paradigm from rigid sub-task policies to highly composable \textit{atomic skills} within a receding-horizon control flow. First, a \textbf{Task Planner} dynamically composes these skills on the fly. By replanning after completing each plan, it swiftly adapts to environmental changes and enables open-ended plan generation, bypassing the scalability limits of predefined sub-task lists. Next, by breaking down traditional sub-tasks into finer-grained actions, the \textbf{Skill Executor} employs skill-specific flow-matching experts built upon a shared VLM backbone. This semantic partitioning allows diverse sub-tasks sharing the same underlying action to reuse a single action expert, maximizing reusability while avoiding capacity interference. Finally, a \textbf{Reflection Critic} provides continuous visual verification to trigger flexible replanning. Operating at this granular level allows the Critic to explicitly manage the complex spatial reasoning essential for robust error recovery.

The Outer Loop drives the automated data curation and evolution of the skill library, eliminating the need for manual data curation. First, a VLM-powered \textbf{Data Curator} processes startup demonstrations and deployment rollouts to autonomously segment and verify trajectory clips with instruction captions. Next, an \textbf{Skill Generator} clusters these instructions to establish a discrete vocabulary of atomic skills, seamlessly standardizing new rollout data into this inventory. Finally, a \textbf{Skill Trainer} uses this curated data to instantiate and continuously fine-tune the skill policies.

To evaluate MobiAgent on long-horizon mobile manipulation, we conduct experiments across the simulated RoboCasa~\citep{robocasa2024,robocasa365} and BEHAVIOR-1K~\citep{li2023behavior1k} benchmarks, as well as four real-world tasks on Astribot S1. On BEHAVIOR-1K, MobiAgent achieves a 65.0\% mean success rate, outperforming the task-specific $\pi_{0.5}$-TA baseline~\citep{larchenko2025behavior1ksolution} by 22.5 percentage points, while component ablations demonstrate the importance of dynamic planning, skill-specific experts, and visual verification. Its critic-driven global replanning further enables recovery from failures such as object drops and pose misalignments. Finally, deployment-to-training updates improve success from 7.50\% to 27.50\% on RoboCasa and from 32.5\% to 57.5\% on Astribot S1, validating the Outer Loop's ability to drive continuous, annotation-free policy improvement through autonomous data recycling.

\section{Related Work}
\label{sec:related_work}

\textbf{Vision-language-action (VLA) policies for short-horizon manipulation.}
Recent foundation models have popularized VLA policies that directly map language and visual observations to short-horizon robot actions \citep{brohan2023rt2, kim2024openvla, octo2024, physicalintelligence2024pi0, physicalintelligence2025pi0_5, li2024cogact, zhen2025spatialvla, kim2025openvlaoft, li2024roboflamingo, zhang2026retrieval}. To model these short-horizon action distributions, architectures typically adopt one of two paradigms: models like RT-2 \citep{brohan2023rt2} and OpenVLA \citep{kim2024openvla} co-train web-scale VLMs to output discretized action tokens, whereas recent state-of-the-art approaches like $\pi_{0}$ \citep{physicalintelligence2024pi0} and $\pi_{0.5}$ \citep{physicalintelligence2025pi0_5} leverage flow matching to directly generate continuous action trajectories.

\textbf{Hierarchical agentic frameworks for long-horizon tasks.}
Hierarchical frameworks typically layer LLMs or VLMs over low-level controllers using a plan-execute-verify paradigm \citep{yao2023react, liang2023codeaspolicies, huang2023voxposer, li2024manipllm, wu2024momallm, shi2025hirobot, zhao2025vla2, pang2026scivla, guo2024doremi, liu2023reflect, duan2024aha, liu2026goal2skill}. While effective, they often rely on fixed skill repertoires. For instance, CaP-X \citep{fu2026capx} composes predefined primitives but cannot learn underlying continuous sensorimotor behaviors. MobiAgent instead employs a unified, skill-conditioned VLA for native vision-based control. Additionally, existing error recovery mechanisms are often rigid; RoboClaw \citep{roboclaw2026}, for example, requires explicitly paired forward and reset policies. Adaptive World Action Models use future--reality verification~\citep{wang2026trust}, whereas MobiAgent performs skill-level verification and task-level replanning.
Finally, unlike most open-loop systems that leave failed behaviors unchanged, MobiAgent introduces a deployment-to-training pipeline to continuously refine its skills from curated rollouts.

\textbf{Mobile manipulation systems.} 
Current mobile manipulation systems \citep{wu2023tidybot, xiong2024spotcompose, fu2024mobilealoha, fu2024humanplus, liu2024okrobot, wang2025mg, yenamandra2023homerobot}, including recent VLA extensions like Mobi-\(\pi\) \citep{yang2025mobipi} and N2M \citep{chai2025n2m}, excel at reactive control and navigation-manipulation hand-offs. However, they lack the closed-loop replanning, online verification, and deployment-driven refinement required for long-horizon execution. Furthermore, state-of-the-art approaches like \(\pi_{0.5}\)-TA \citep{larchenko2025behavior1ksolution} (the 2025 BEHAVIOR Challenge winner) rely on isolated, task-specific models. Their use of time-indexed stages forces distinct behaviors to share a single prediction head, causing capacity interference.

\section{MobiAgent: A Dual-Loop Agentic System for Mobile Manipulation}
\label{sec:method}

\begin{figure}[t]
\centering
\includegraphics[width=0.97\linewidth]{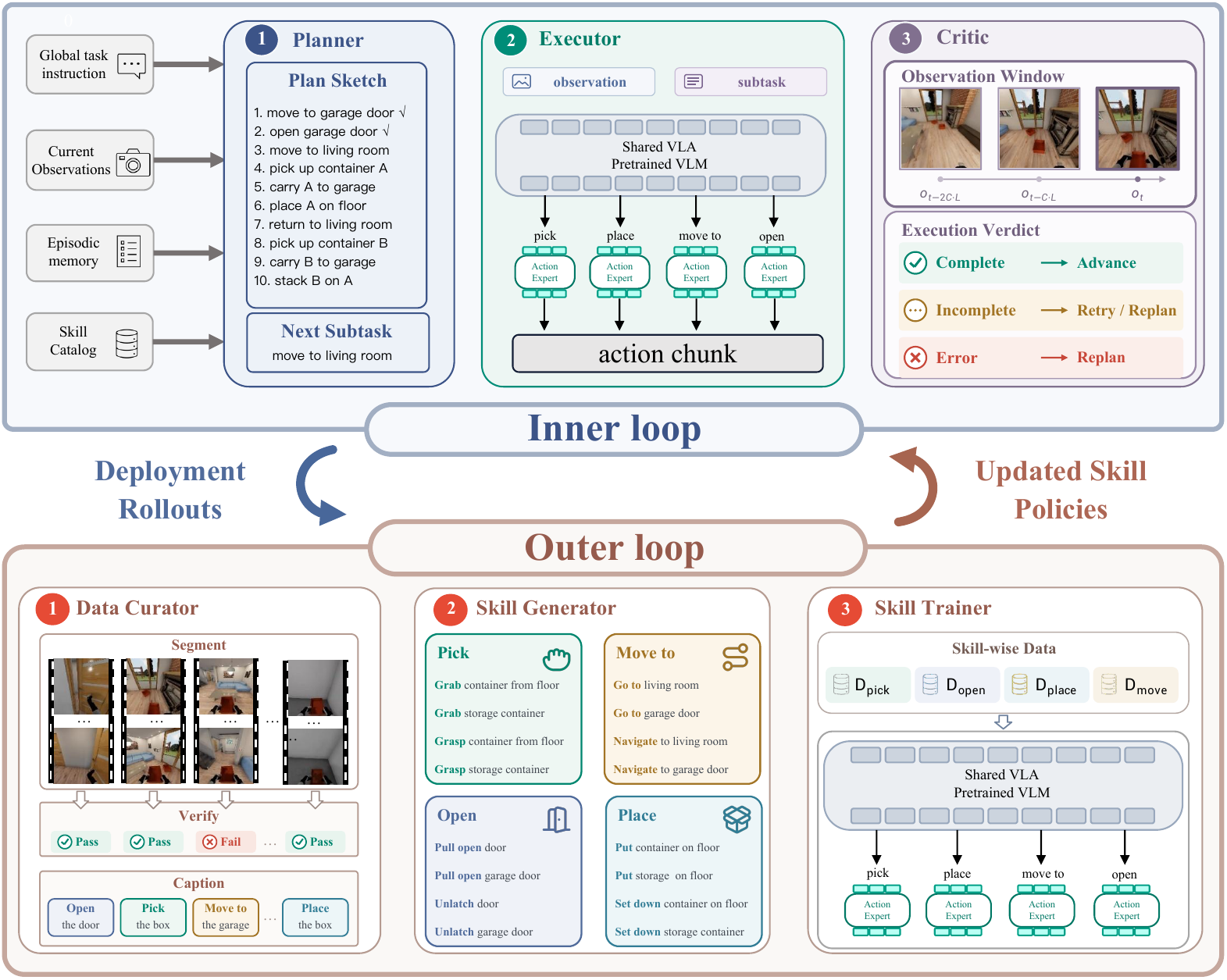}
\vspace{-0.1cm}
    \caption{\textbf{Illustration of the MobiAgent framework.} 
    A dual-loop agentic system bridging robust execution and continuous policy evolution. (a) The \textbf{Inner Loop} ensures adaptive deployment via a Task Planner for dynamic skill composition, a Skill Executor utilizing reusable skill-specific policies, and a Reflection Critic that triggers flexible replanning for robust error recovery. (b) The \textbf{Outer Loop} drives automated self-improvement via a Data Curator, Skill Generator, and Skill Trainer, which autonomously extract, standardize, and fine-tune atomic skills from trajectory data.
    }
\label{fig:method_overview}
\end{figure}

\textbf{Task Definition \& Motivation.} We address long-horizon mobile manipulation, where a robot must execute complex, multi-stage tasks specified by a global language instruction \(\ell\). At each step \(t\), the robot maps multi-modal observations \(o_t\) to continuous base and manipulator actions \(\mathbf{a}_t\), requiring the seamless interleaving of distinct behavioral modes like global navigation and precise prehension. Since monolithic end-to-end policies often suffer from compounding execution errors and capacity interference when jointly modeling disparate sensorimotor mappings, we propose \textbf{MobiAgent}. By decoupling high-level semantic reasoning from low-level sensorimotor control, our system enables dynamic task decomposition, spatial-aware failure recovery, and continuous offline skill evolution.

\textbf{Dual-Loop Architecture.} As shown in Fig.~\ref{fig:method_overview}, MobiAgent employs a dual-loop agentic architecture to separate deployment-time execution from offline policy evolution:
\begin{itemize}
    \item \textbf{Inner Loop (Deployment):} Governs closed-loop execution. It consists of a \textbf{Task Planner} that dynamically decomposes \(\ell\) into actionable sub-tasks, a \textbf{Skill Executor} that maps sub-task instructions and real-time observations to low-level action chunks, and a \textbf{Reflection Critic} that evaluates outcomes to trigger progression, retries, or spatial-aware replanning.
    \item \textbf{Outer Loop (Offline Training):} Drives automated policy evolution. It ingests collected rollouts, utilizing a \textbf{Data Curator} for VLM-powered trajectory segmentation, a \textbf{Skill Generator} for semantic skill clustering, and a \textbf{Skill Trainer} for continuous policy updates. 
\end{itemize}

\subsection{The Inner Loop: Closed-Loop Agentic Execution}
\label{sec:sys_inner_loop}

The inner loop performs closed-loop task execution at deployment time. As shown in Fig.~\ref{fig:method_overview}, it repeatedly plans the next sub-task, executes a skill-conditioned action chunk, and evaluates the result. This process continues until the task succeeds or a fixed budget is exhausted. The overall control flow is not an open-loop decomposition followed by blind execution; instead, it is a recurrent plan--execute--verify loop grounded in current observations and recent failures.

\subsubsection{Task Planner: Dynamic Atomic Skill Composition}

The Task Planner, instantiated as a Vision-Language Model (VLM) \(\Pi_{\text{Plan}}\), shifts the execution paradigm from rigid sub-task mapping to the dynamic composition of composable \textit{atomic skills}. Crucially, rather than generating a rigid, full-horizon plan from a predefined list of high-level sub-tasks, \(\Pi_{\text{Plan}}\) operates in an open-ended, receding-horizon manner. It is invoked on the fly at specific decision steps \(k\), triggered either by the successful completion of a preceding skill or by an execution failure. This formulation bypasses the scalability limits of static planning, allowing the agent to dynamically compose fine-grained skills to swiftly adapt to environmental changes and perturbations.

At each invocation step \(k\), the Task Planner emits the next active sub-task conditioned on four inputs:
\[
\ell_k,\, s_k \,=\, \Pi_{\text{Plan}}\!\left(\ell,\, o_k,\, \mathcal{H}_k,\, \mathcal{S}\right).
\]
Here, \(\ell\) is the global task instruction, and \(o_k\) denotes the current multi-camera observation. \(\mathcal{H}_k\) represents the episode memory, a running log of past sub-tasks, verification verdicts, and observation cues (e.g., spatial misalignments) generated by the Reflection Critic. \(\mathcal{S}\) is the atomic skill catalog, exposed as an open-vocabulary registry. For the outputs, \(\ell_k\) is a compact natural-language instruction for the immediate sub-task (e.g., ``pick the red can''), and \(s_k \in \mathcal{S}\) is a skill hint (e.g., \texttt{pick}). This hint serves as the critical interface for the Skill Executor to select the corresponding policy.

\subsubsection{Skill Executor: Skill-Specific Action Generation}
\label{sec:method_executor}

The Skill Executor, denoted as \(\Pi_{\text{Exec}} = \{\Pi_{\text{Exec}}^{(s_k)}\}_{s_k \in \mathcal{S}}\), operates as a modular library of skill-specific policies dynamically indexed by the hint \(s_k\). At each low-level control step \(t\) within the \(k\)-th sub-task, the Executor routes \(\ell_k\) and the current multi-modal observation \(o_t\) to the selected policy, predicting a short-horizon action chunk of length \(L\):
\[
\mathbf{a}_{t:t+L} \,=\, \Pi_{\text{Exec}}^{(s_k)}\!\left(\ell_k,\, o_t\right).
\]
This semantic partitioning allows the system to specialize policies for distinct behavioral modes (e.g., locomotion vs. arm control) without suffering from the capacity interference.

We instantiate \(\Pi_{\text{Exec}}\) using the \(\pi_{0.5}\) Vision-Language-Action (VLA) model~\citep{physicalintelligence2025pi0_5}. To maximize reusability while preserving optimization efficiency, we introduce a decoupling strategy: maintaining a shared VLM trunk \(\pi_{\text{vlm}}\) across all skills while instantiating independent flow-matching action experts \(\pi_{\text{action}}^{(s_k)}\) for each skill. This is formulated as a function composition:
\[
\mathbf{a}_{t:t+L} \,=\, \Pi_{\text{Exec}}^{(s_k)}\!\left(\ell_k,\, o_t\right) \,=\, \pi_{\text{action}}^{(s_k)}\!\left( \pi_{\text{vlm}}\!\left(\ell_k,\, o_t\right) \right).
\]

\subsubsection{Reflection Critic: Continuous Evaluation and Feedback}

To prevent compounding errors during long-horizon execution, we introduce a closed-loop feedback mechanism driven by the Reflection Critic \(\Pi_{\text{Critic}}\), which  evaluates the execution outcomes to decide whether the active sub-task has been successfully completed, remains incomplete, or has failed. 

Specifically, after the execution of every \(C\) action chunks (totaling \(C \cdot L\) actions), \(\Pi_{\text{Critic}}\) processes \(\ell_k\) alongside a stride-sampled observation window \(x_t = \left(o_{t-2C\cdot L},\,o_{t-C\cdot L},\,o_t\right)\). Relying exclusively on deployable observations, the Critic maps these inputs to a structured decision:
\[
v_t,\, \phi_t,\, c_t = \Pi_{\text{Critic}}\!\left( \ell_k, \, x_t \right) .
\]
The decision comprises a status verdict \(v_t \in \{\texttt{complete},\, \texttt{incomplete},\, \texttt{error}\}\), a follow-up recommendation \(\phi_t \in \{\texttt{next},\, \texttt{retry},\, \texttt{replan}\}\), and a set of observable cues \(c_t\) for offline auditing. To ensure deterministic control, the outputs are strictly constrained to a legal joint schema: 1) \(v_t=\texttt{complete} \Rightarrow \phi_t=\texttt{next}\); 2) \(v_t=\texttt{incomplete} \Rightarrow \phi_t \in \{\texttt{retry},\, \texttt{replan}\}\); and 3) \(v_t=\texttt{error} \Rightarrow \phi_t=\texttt{replan}\). Any schema violation is rejected, and \(v_t\) is set to \texttt{error}.

An orchestration controller \(\Pi_{\text{Orch}}\) governs the discrete state transitions by applying a deterministic policy over the validated \(\left(v_t,\, \phi_t\right)\) tuples. Conditioned on a per-sub-task evaluation counter \(n_t\) and an upper bound \(n_\text{max}\), the transition dynamics are defined by the following mutually exclusive rules:
\[
\Pi_{\text{Orch}}\left(v_t,\, \phi_t,\,n_t\right) =
\begin{cases}
\textsc{advance}, & v_t=\texttt{complete}, \\
\textsc{replan}, & v_t=\texttt{error} \lor \phi_t=\texttt{replan}, \\
\textsc{retry}, & v_t=\texttt{incomplete} \land \phi_t=\texttt{retry} \land n_t < n_\text{max}, \\
\textsc{replan}, & v_t=\texttt{incomplete} \land \phi_t=\texttt{retry} \land n_t \ge n_\text{max}.
\end{cases}
\]
Operationally, a \textsc{retry} command triggers continued execution of the active sub-task. Conversely, both \textsc{advance} and \textsc{replan} halt the low-level execution and yield control back to the Planner. To facilitate spatial-aware replanning during this phase, the Critic's evaluation logs and spatial cues \(c_t\) are persistently appended to the episodic memory \(\mathcal{H}_k\).

\subsection{The Outer Loop: Automated Agentic Training}
\label{sec:sys_outer_loop}

The outer loop governs the offline training and continuous evolution of the skill library. It operates in two primary phases: first, \textit{bootstrapping}, which initializes the system using domain-specific startup data (e.g., synthetic trajectories for simulation benchmarks or human-teleoperated demonstrations for real-world applications); and second, \textit{continuous finetuning}, which runs offline after episode termination to curate deployment rollouts and iteratively refine the active policies. 

To convert both startup data and deployment rollouts into actionable updates, trajectories are automatically processed via a three-stage pipeline (Fig.~\ref{fig:method_overview}): 1) the \textbf{Data Curator} segments long trajectories into candidate clips paired with instruction captions, serving as atomic skill demonstrations; 2) the \textbf{Skill Generator} aggregates these captioned clips to formalize a discrete set of canonical atomic skills; and 3) the \textbf{Skill Trainer} utilizes the curated data to instantiate or update skill-wise policies.

\subsubsection{Data Curator: VLM-Powered Trajectory Segmenting}

The Data Curator transforms raw trajectories into candidate training clips with meaningful sub-task boundaries via an offline \textbf{segment-and-verify} pipeline (Fig.~\ref{fig:method_overview}). 

\textbf{Segmenting}. A video-capable VLM processes sampled frames from the full trajectory to propose temporal boundaries aligned with stage-level behaviors. For each segmented clip, the VLM generates a free-form action description. To ensure precise video clip captioning, the VLM employs a hierarchical reasoning process: it first determines whether the segment entails pure locomotion (without manipulation) and, if not, further infers the specific semantics of the manipulation types.

\textbf{Verification}. A separate VLM evaluates candidate clips against their proposed descriptions using a rigorous quality-control prompt. This verification process explicitly assesses temporal continuity, logical skill adjacency, and object-instruction consistency. Segments that accurately realize the intended behavior are accepted, whereas low-quality or boundary-misaligned clips are rejected and iteratively routed back for boundary refinement or instruction correction until they pass verification.

\subsubsection{Skill Generator: Atomic Skill Clustering}

Following trajectory curation, the Skill Generator standardizes the diverse, free-form instructions of verified clips into a compact, reusable inventory of atomic skills. For \textit{startup data}, the system constructs the skill vocabulary from scratch through a three-step pipeline: first, an LLM performs semantic clustering on the instructions, dynamically determining the optimal number of clusters without predefined human priors; second, it assigns a canonical name to each cluster based on frequent textual patterns, enforced by a rigorous coverage check to resolve any omissions; third, the LLM standardizes the original instructions by replacing diverse action verbs with these newly unified canonical names. For \textit{rollout data}, the clustering and naming phases are bypassed; the LLM directly assigns the new clips to the previously established atomic skills and modifies their instructions accordingly. Ultimately, this adaptive approach transforms raw, heterogeneous fragments into a controlled skill vocabulary, providing a structured data substrate for downstream policy training.

\subsubsection{Skill Trainer: Continuous Policy Evolution}

The Skill Trainer completes the outer loop by translating the structured data into executable neural policies. During the bootstrapping phase with startup data, the Trainer dynamically instantiates dedicated action experts corresponding to the optimal number of canonical skills identified by the Generator. Sharing a common VLM trunk initialized from a pre-trained \(\pi_{0.5}\) model (Sec.~\ref{sec:method_executor}), these experts are subsequently trained via behavior cloning on their respective data clusters to establish the initial skill library. Conversely, during the continuous finetuning phase, the Trainer leverages newly curated deployment rollouts to directly update the existing policies, progressively refining their closed-loop robustness without expanding the model inventory.

\section{Experiments}
\label{sec:experiments}

\subsection{Main Results}
\label{sec:main_results}

We evaluate MobiAgent in three settings, including two simulation benchmarks, i.e., RoboCasa~\citep{robocasa2024,robocasa365} and BEHAVIOR-1K~\citep{li2023behavior1k}, and real-world mobile manipulation. RoboCasa evaluates multi-round deployment-to-training improvement using a Franka arm mounted on an Omron mobile base. BEHAVIOR-1K evaluates long-horizon performance using a Galaxea R1 Pro. The real-world experiments is conducted on Astribot S1.

\noindent\textbf{RoboCasa.}
We evaluate on RoboCasa365 Composite-Seen, which contains 16 household tasks comprising two to fifteen subtasks. Using a Franka arm mounted on an Omron mobile base, we conduct five trials per target scene and average success across tasks and scenes. CaP-X~\citep{fu2026capx} and the base $\pi_{0.5}$ policy~\citep{physicalintelligence2025pi0_5} achieve 5.00\% and 7.50\%, respectively. MobiAgent improves from 18.75\% in Round~1 to 27.50\% in Round~3 through successive deployment-to-training updates, before stabilizing at 25.00--27.50\% in Rounds~4--6. The early gains demonstrate the benefit of converting deployment rollouts into skill-wise training data, while the later saturation indicates the need for stronger atomic policies and more diverse experience.

\begin{table}[t]
    \centering
    \caption{
        Success rates (\%) on RoboCasa Composite-Seen \citep{robocasa2024,robocasa365}.
        R0 denotes the base $\pi_{0.5}$ policy, while R1--R6 denote
        successive deployment-to-training rounds of MobiAgent.
    }
    \label{tab:robocasa_main}
    \resizebox{0.85\linewidth}{!}{%
    \begin{tabular}{lcccccccc}
        \toprule
        Method
        & \multicolumn{2}{c}{Baselines}
        & \multicolumn{6}{c}{MobiAgent} \\
        \cmidrule(lr){2-3}
        \cmidrule(lr){4-9}
        & CaP-X \cite{fu2026capx}
        & $\pi_{0.5}$ \cite{physicalintelligence2025pi0_5} (R0)
        & R1
        & R2
        & R3
        & R4
        & R5
        & R6 \\
        \midrule
        Success Rate (\%)
        & 5.00
        & 7.50
        & 18.75
        & 21.25
        & \textbf{27.50}
        & 25.00
        & \textbf{27.50}
        & 26.25 \\
        \bottomrule
    \end{tabular}}
\end{table}

\begin{table}[t]
    \centering
    \caption{
        Success rates (\%) on four BEHAVIOR-1K tasks.
        MobiAgent shares skill policies across tasks, whereas
        $\pi_{0.5}$-TA uses a separately trained model for each task.
    }
    \label{tab:behavior_main}
    \resizebox{0.85\linewidth}{!}{%
    \begin{tabular}{lccccc}
        \toprule
        Method
            & \texttt{push-radio}
            & \texttt{dispose-trash}
            & \texttt{stack-storage}
            & \texttt{fetch-beer}
            & Mean \\
        \midrule
        $\pi_{0.5}$-TA~\citep{larchenko2025behavior1ksolution}
            & 60 & 40 & 60 & 10 & 42.5 \\
        MobiAgent (Ours)
            & \textbf{70}
            & \textbf{60}
            & \textbf{70}
            & \textbf{60}
            & \textbf{65.0} \\
        \bottomrule
    \end{tabular}%
    }
\end{table}

\noindent\textbf{BEHAVIOR-1K.}
We evaluate four long-horizon tasks in OmniGibson~\citep{li2022igibson2}: \texttt{push-radio}, \texttt{dispose-trash}, \texttt{stack-storage}, and \texttt{fetch-beer}, covering articulated interaction, cross-room transport, stacking, and repeated retrieval. We compare against $\pi_{0.5}$-TA~\citep{larchenko2025behavior1ksolution}, the first-place solution of the 2025 BEHAVIOR Challenge, which trains a separate model for each task. In contrast, MobiAgent shares reusable skill policies across tasks and composes them through closed-loop planning. Under matched conditions with 10 trials per task, MobiAgent achieves 65.0\% mean success, outperforming $\pi_{0.5}$-TA's 42.5\% by 22.5 percentage points (\Cref{tab:behavior_main}). The gain reaches 50 points on the longest \texttt{fetch-beer} task, highlighting the advantage of skill reuse and closed-loop replanning.

\begin{figure}[t]
 \centering
 \newlength{\realimgheight}
 \setlength{\realimgheight}{0.13\textwidth}
 \begin{subfigure}{0.19\linewidth}
  \centering
  \includegraphics[width=\linewidth,height=\realimgheight,keepaspectratio]{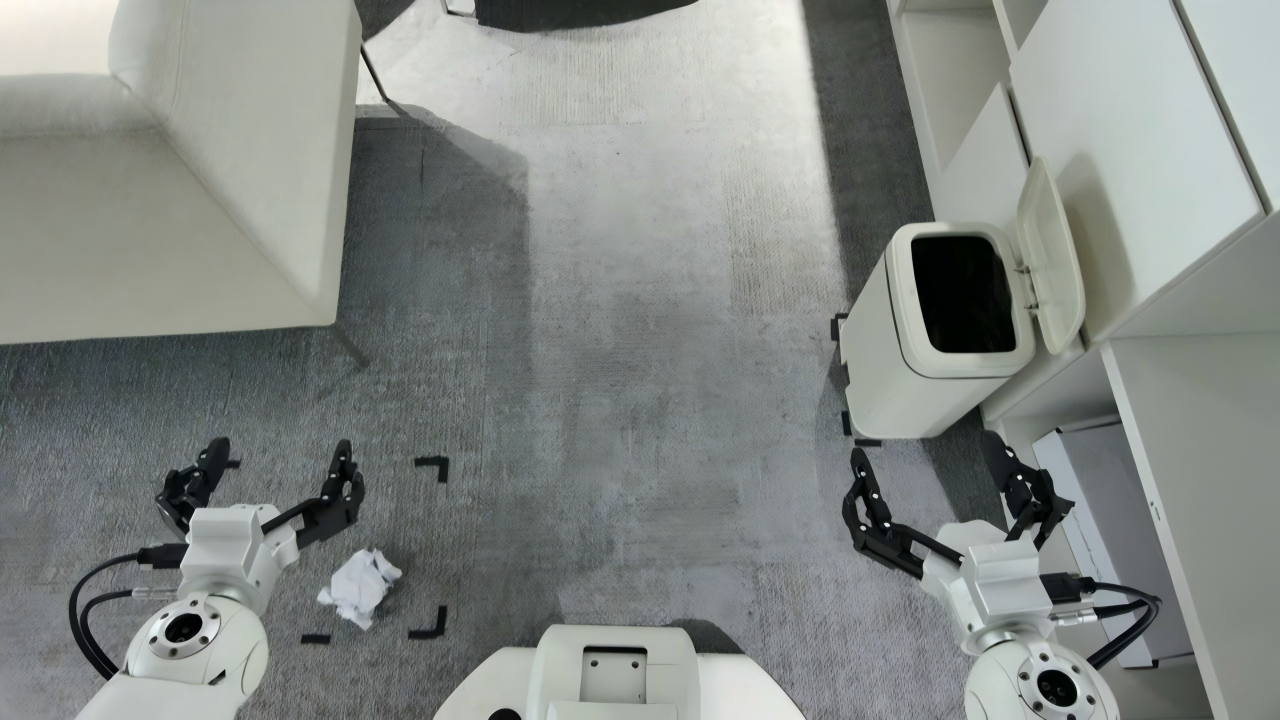}
  \caption{\texttt{trash-garbage}}
 \end{subfigure}\hfill
 \begin{subfigure}{0.19\linewidth}
  \centering
  \includegraphics[width=\linewidth,height=\realimgheight,keepaspectratio]{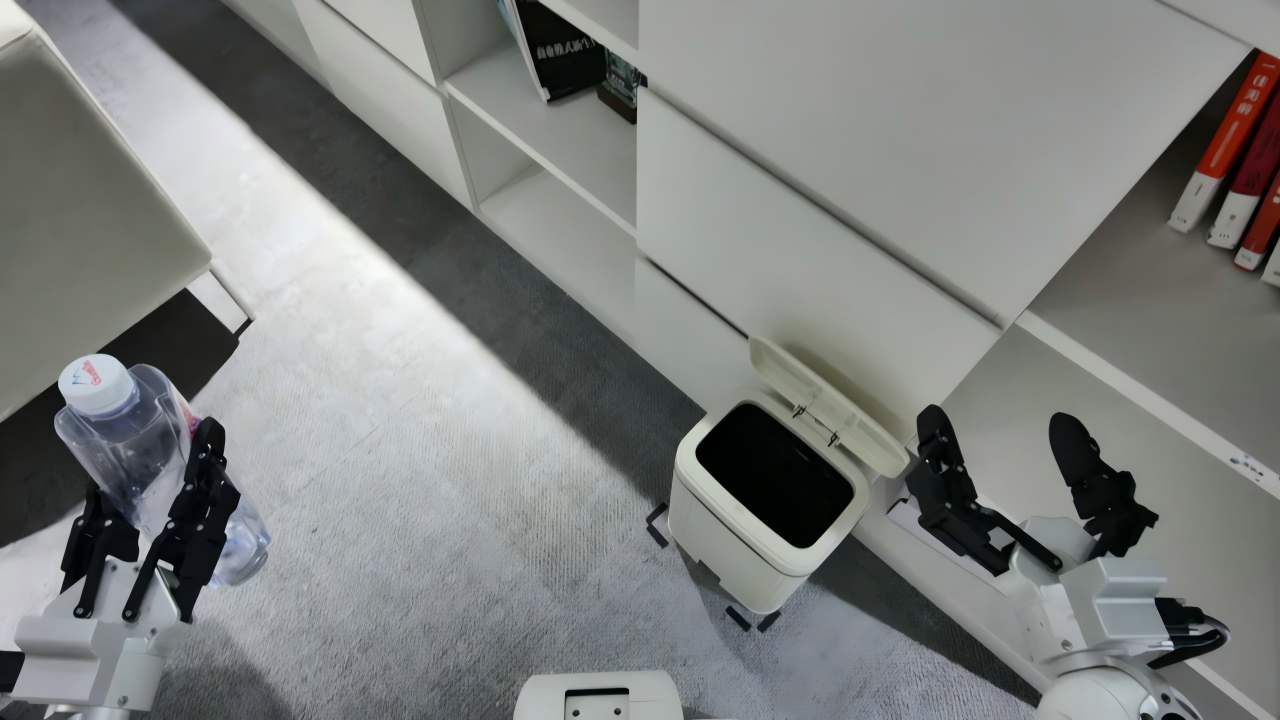}
  \caption{\texttt{trash-bottle}}
 \end{subfigure}\hfill
 \begin{subfigure}{0.19\linewidth}
  \centering
  \includegraphics[width=\linewidth,height=\realimgheight,keepaspectratio]{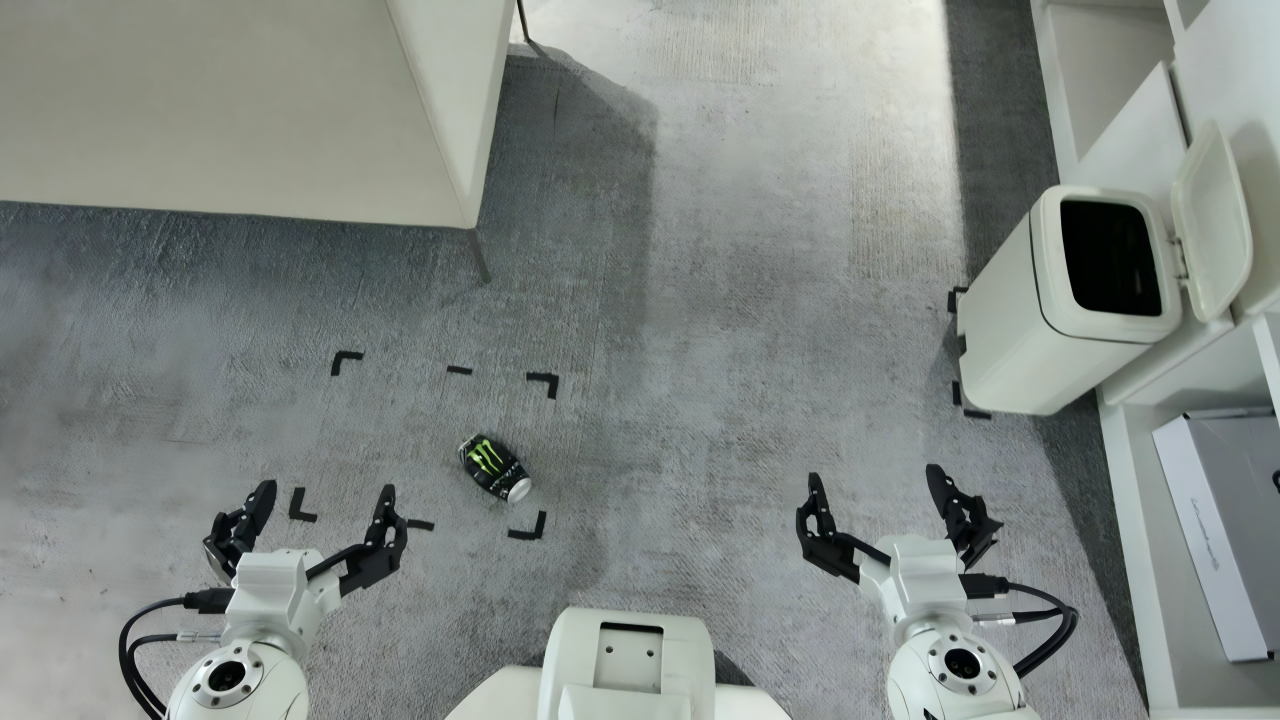}
  \caption{\texttt{trash-can}}
 \end{subfigure}\hfill
 \begin{subfigure}{0.19\linewidth}
  \centering
  \includegraphics[width=\linewidth,height=\realimgheight,keepaspectratio]{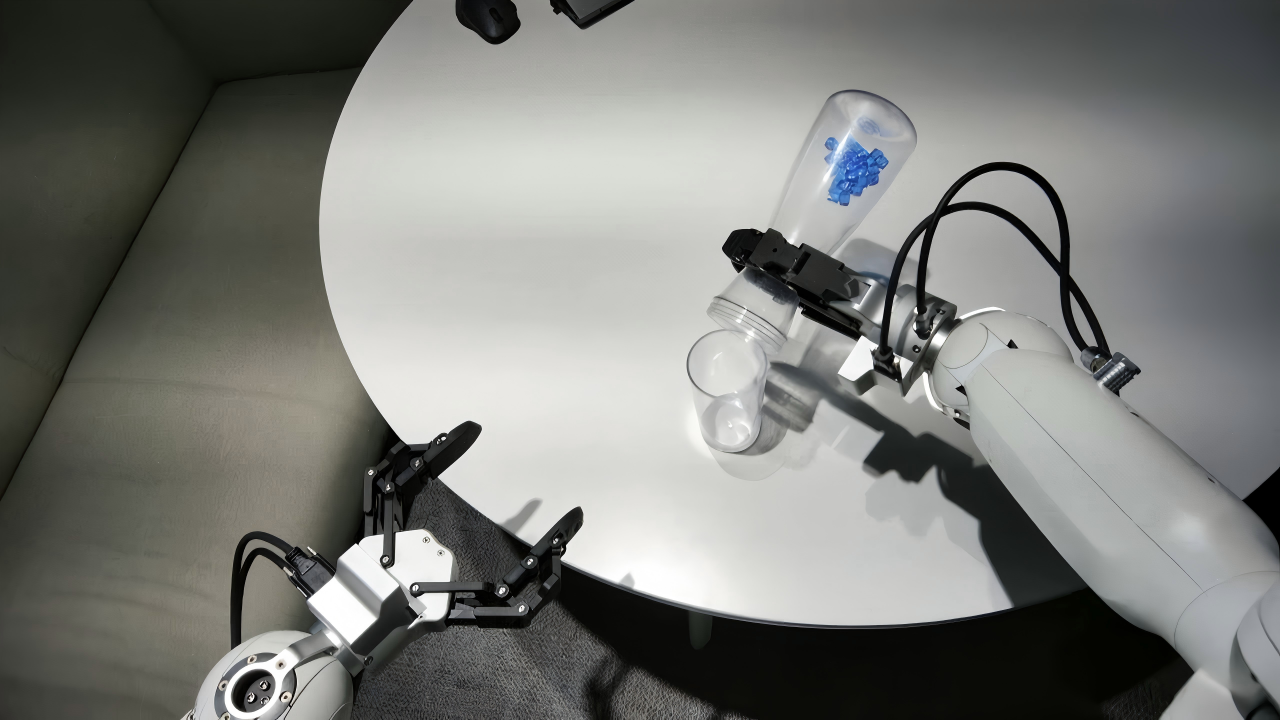}
  \caption{\texttt{pour-blue}}
 \end{subfigure}\hfill
 \begin{subfigure}{0.19\linewidth}
  \centering
  \includegraphics[width=\linewidth,height=\realimgheight,keepaspectratio]{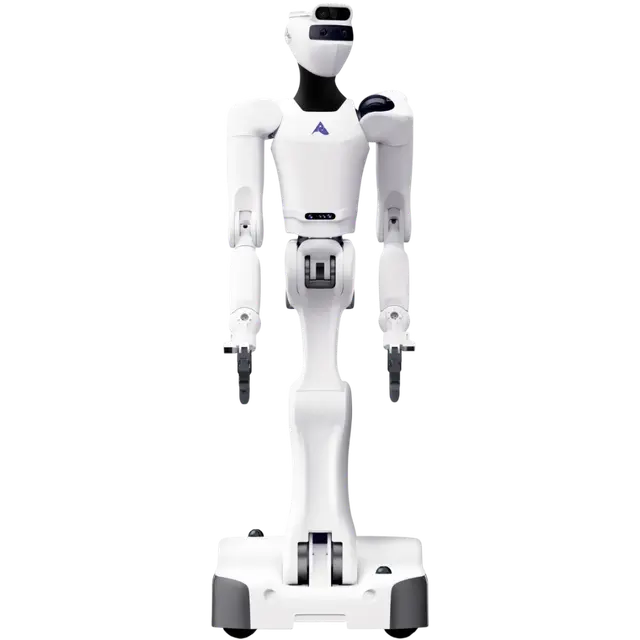}
  \caption{Astribot S1}
 \end{subfigure}
 \caption{Representative real-robot evaluation scenes and the Astribot S1 platform.}
 \label{fig:real_tasks}
\end{figure}

\noindent \textbf{Real-world experiments.}
To evaluate the Outer Loop's capacity for automated policy evolution, we deploy MobiAgent on Astribot S1~\citep{gao2025astribot}, a bimanual mobile manipulator equipped with multi-view cameras and proprioceptive sensors operating at 30\,Hz. We evaluate four long-horizon tasks that share skill-specific policies (Fig.~\ref{fig:real_tasks}): disposing of garbage, a bottle, and a can into a trash can (\texttt{trash-garbage}, \texttt{trash-bottle}, and \texttt{trash-can}), and pouring blue particles into a cup (\texttt{pour-blue}). Using 100 demonstrations per task, the monolithic $\pi_{0.5}$ baseline~\citep{physicalintelligence2025pi0_5} achieves averaged success rates of 10.0\%. In comparison, MobiAgent reaches 32.5\% in the bootstrap round. After each deployment round, the Outer Loop automatically segments and verifies successful rollouts, adding approximately 20 trajectories per task to refine the corresponding skill policies without manual annotation. As shown in \Cref{tab:self_improve}, the mean success rate further increases to 40.0\% and 57.5\% after two updates, demonstrating both the benefit of skill-based decomposition over the monolithic baseline and continuous policy improvement through automated deployment-data recycling.

\begin{table}[t]
    \centering
    \caption{
        Success rates (\%) on four real-world tasks.
        Both $\pi_{0.5}$ and MobiAgent Round~1 use 100 demonstrations per task.
        Subsequent rounds refine MobiAgent using approximately 20 automatically
        curated deployment trajectories per task.
    }
    \label{tab:self_improve}
    \resizebox{0.9\linewidth}{!}{%
    \begin{tabular}{lccccc}
        \toprule
        Method / Round
        & \texttt{trash-garbage}
        & \texttt{trash-bottle}
        & \texttt{trash-can}
        & \texttt{pour-blue}
        & Mean \\
        \midrule
        $\pi_{0.5}$~\citep{physicalintelligence2025pi0_5}
        & 10.0 & 20.0 & 10.0 & 0.0 & 10.0 \\
        \midrule
        MobiAgent R1 (Bootstrap)
        & 40.0 & 50.0 & 20.0 & 20.0 & 32.5 \\
        MobiAgent R2
        & 40.0 & 60.0 & 30.0 & 30.0 & 40.0 \\
        MobiAgent R3
        & \textbf{50.0} & \textbf{60.0} & \textbf{70.0}
        & \textbf{50.0} & \textbf{57.5} \\
        \bottomrule
    \end{tabular}%
    }
\end{table}

\subsection{Ablation Studies}
\label{sec:ablations}

We ablate the principal design choices of MobiAgent on all four BEHAVIOR-1K tasks. As summarized in \Cref{tab:ablations}, each component contributes substantially to the full system's performance:

\noindent \textbf{Task Planner and shared atomic skills.}
The fixed sub-task baseline does not perform planning. Each task follows a predefined sub-task sequence, and every task-specific sub-task is executed by a separately trained model. Thus, semantically identical behaviors are not shared across tasks; for example, the \texttt{pick} sub-tasks in different tasks use different policies. MobiAgent instead composes task-agnostic atomic skills, allowing all \texttt{pick} sub-tasks to reuse the same \texttt{pick} policy. Replacing this design with fixed, task-specific sub-task models reduces mean success from 65.0\% to 20.0\%.

\noindent \textbf{Skill-specific experts.}
We replace the skill-specific action experts with a single action head while retaining the shared VLM backbone, Planner, Critic, and training data. The mean success rate decreases from 65.0\% to 42.5\%, indicating that specialized action experts mitigate interference among navigation, grasping, placement, and articulated-object interaction.

\noindent \textbf{Reflection Critic.}
Removing the Reflection Critic causes the largest degradation, reducing mean success from 65.0\% to 10.0\%. Without visual outcome verification, the system cannot reliably determine whether an active skill has completed or failed, causing execution errors to propagate across subsequent steps.

\begin{table}[t]
    \centering
    \caption{
        Component ablations on BEHAVIOR-1K.
        All entries report task success rates (\%).
    }
    \label{tab:ablations}
    \resizebox{0.98\linewidth}{!}{%
    \begin{tabular}{lccccc}
        \toprule
        Method
        & \texttt{push-radio}
        & \texttt{dispose-trash}
        & \texttt{stack-storage}
        & \texttt{fetch-beer}
        & Mean \\
        \midrule
        Fixed task-specific sub-tasks (w/o Planner)
        & 50 & 0 & 0 & 30 & 20.0 \\
        Single shared action head
        & 40 & 40 & 50 & 40 & 42.5 \\
        w/o Reflection Critic
        & 30 & 10 & 0 & 0 & 10.0 \\
        w/o Global Replanning
        & 70 & 40 & 30 & 20 & 40.0 \\
        \midrule
        MobiAgent (Full)
        & \textbf{70}
        & \textbf{60}
        & \textbf{70}
        & \textbf{60}
        & \textbf{65.0} \\
        \bottomrule
    \end{tabular}%
    }
\end{table}

\subsection{Robustness and Error Recovery}
\label{sec:robustness}

\begin{figure}[t]

 \centering

 \begin{minipage}[t]{0.49\linewidth}
  \centering
  \begin{subfigure}{0.24\linewidth}
   \centering
   \includegraphics[width=\linewidth]{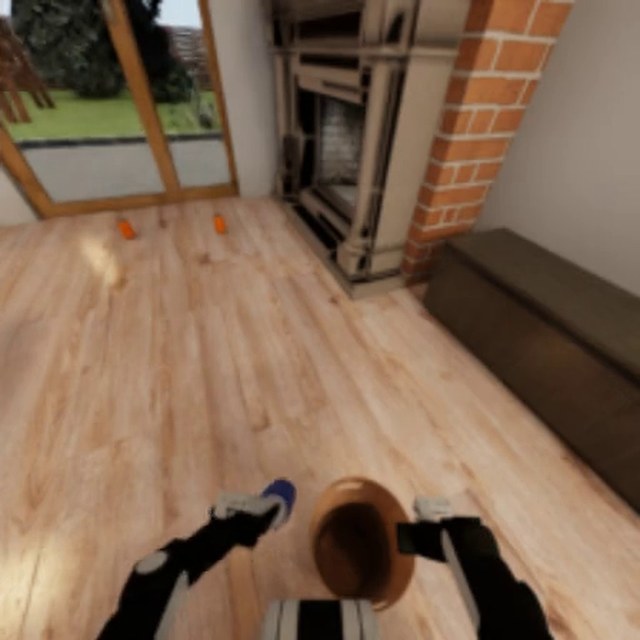}
   \caption*{\scriptsize \textbf{(a)} \texttt{pick\_up}\,\cmark}
  \end{subfigure}\hfill
  \begin{subfigure}{0.24\linewidth}
   \centering
   \includegraphics[width=\linewidth]{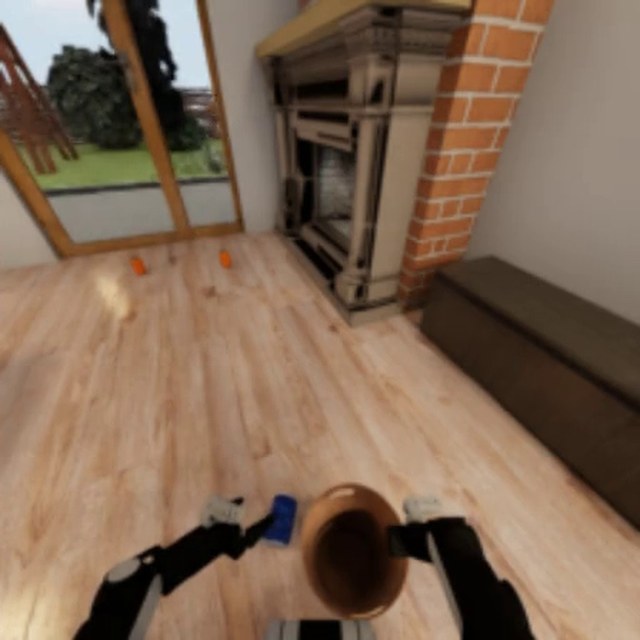}
   \caption*{\scriptsize \texttt{place\_in}\,\xmark}
  \end{subfigure}\hfill
  \begin{subfigure}{0.24\linewidth}
   \centering
   \includegraphics[width=\linewidth]{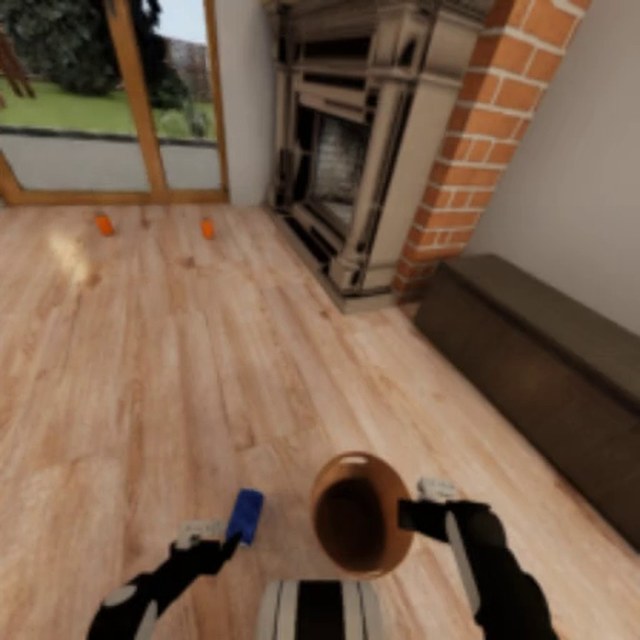}
   \caption*{\scriptsize re-\texttt{move\_to}\,\cmark}
  \end{subfigure}\hfill
  \begin{subfigure}{0.24\linewidth}
   \centering
   \includegraphics[width=\linewidth]{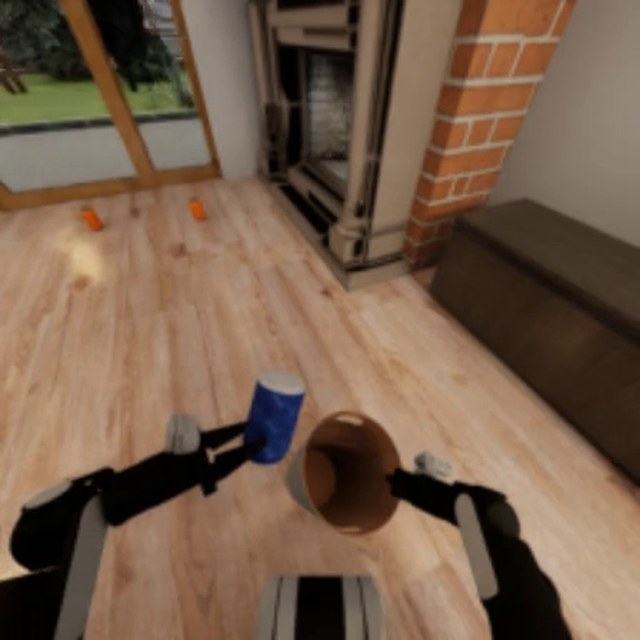}
   \caption*{\scriptsize re-\texttt{pick\_up}\,\cmark}
  \end{subfigure}
 \end{minipage}
 \hfill
 \begin{minipage}[t]{0.49\linewidth}
  \centering
  \begin{subfigure}{0.24\linewidth}
   \centering
   \includegraphics[width=\linewidth]{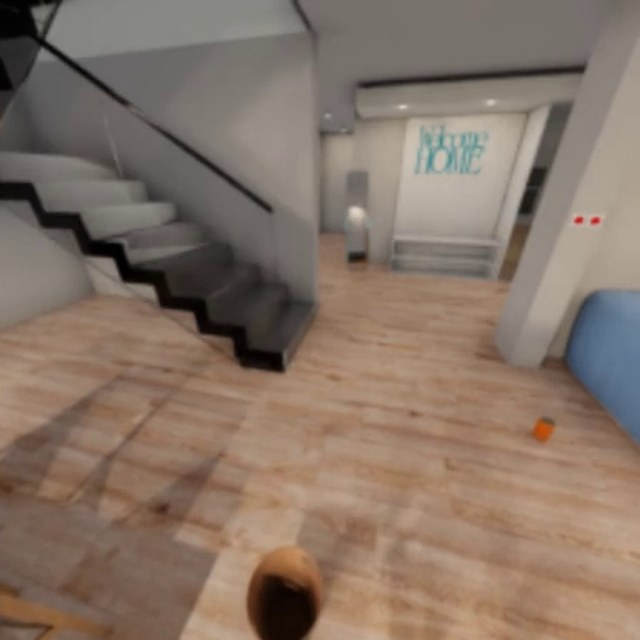}
   \caption*{\scriptsize \textbf{(b)} \texttt{move\_to}\,\cmark}
  \end{subfigure}\hfill
  \begin{subfigure}{0.24\linewidth}
   \centering
   \includegraphics[width=\linewidth]{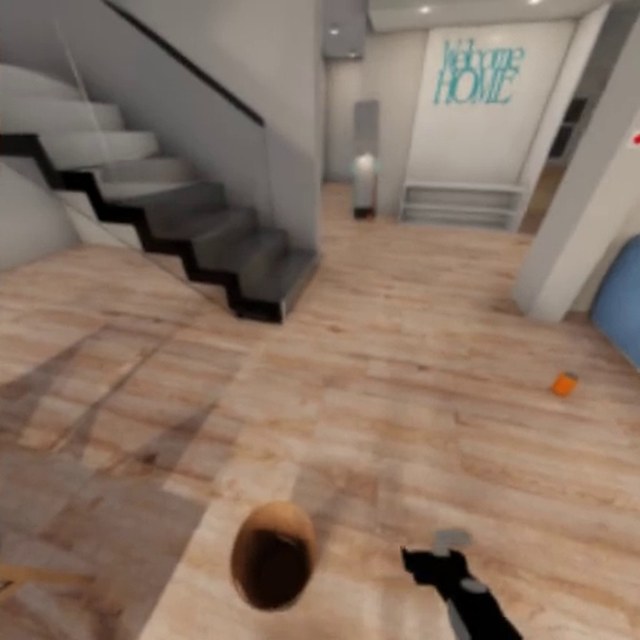}
   \caption*{\scriptsize \texttt{pick\_up}\,\xmark}
  \end{subfigure}\hfill
  \begin{subfigure}{0.24\linewidth}
   \centering
   \includegraphics[width=\linewidth]{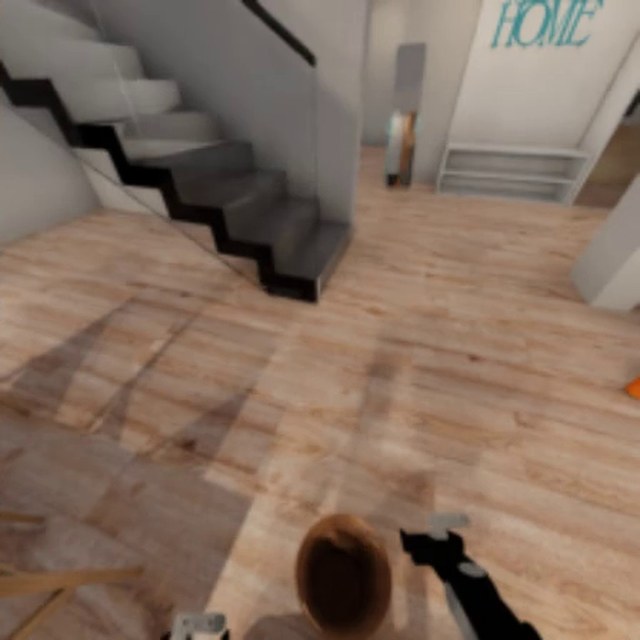}
   \caption*{\scriptsize re-\texttt{move\_to}\,\cmark}
  \end{subfigure}\hfill
  \begin{subfigure}{0.24\linewidth}
   \centering
   \includegraphics[width=\linewidth]{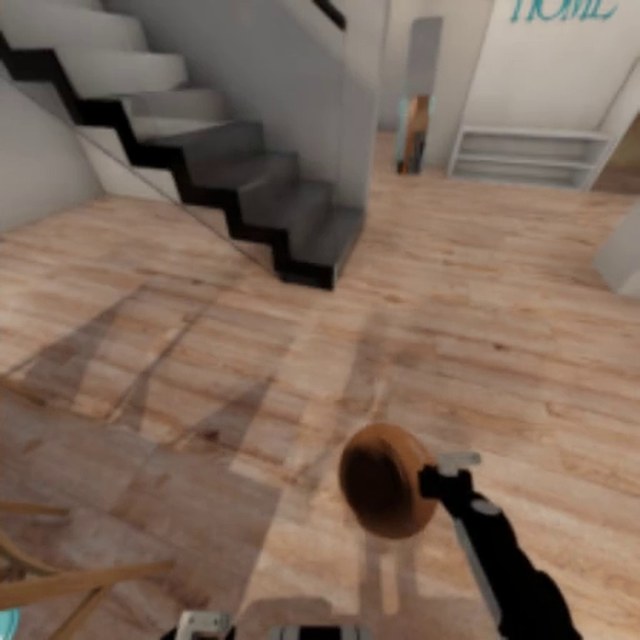}
   \caption*{\scriptsize \texttt{pick\_up}\,\cmark}
  \end{subfigure}
 \end{minipage}
 \caption{\textbf{Examples of robust error recovery.}
 MobiAgent overcomes execution failures through critic-driven global replanning:
 \textbf{(a)} generating new retrieval sub-tasks after an object is dropped, and
 \textbf{(b)} correcting the robot's base pose before retrying the grasp.}
 \label{fig:replan_examples}
\end{figure}

To evaluate MobiAgent's ability to recover from execution failures, we isolate the global replanning mechanism driven by the Reflection Critic. The \textit{w/o Global Replanning} variant retains visual verification but ignores \texttt{REPLAN} signals, restricting the system to local retries of the active skill. As shown in \Cref{tab:ablations}, this variant achieves a mean success rate of 40.0\%, compared with 65.0\% for the full system. On \texttt{dispose-trash}, enabling global replanning improves success from 40\% to 60\%. Across the BEHAVIOR-1K evaluations, 39\% of successful episodes require at least one replan, with successful episodes invoking 0.95 replans on average under a maximum budget of three. As illustrated in \Cref{fig:replan_examples}, MobiAgent generates new retrieval steps after an object is dropped and corrects its base pose before reattempting a failed grasp. Unlike a local retry, which repeats the active skill from the current state, global replanning updates the skill sequence according to the observed post-failure scene, enabling recovery from failures that cannot be resolved by repeating the same low-level behavior.

\section{Conclusion and Limitations}
\label{sec:conclusion}

In this work, we presented MobiAgent, a dual-loop agentic framework for long-horizon mobile manipulation. The Inner Loop replaces rigid sub-task mapping with dynamic composition of atomic skills, enabling receding-horizon execution and critic-driven recovery, while the Outer Loop supports continuous data curation and annotation-free policy self-improvement. Experiments on RoboCasa, BEHAVIOR-1K, and real-world mobile manipulation demonstrate its effectiveness. MobiAgent outperforms the task-specific $\pi_{0.5}$-TA baseline by 22.5 percentage points on BEHAVIOR-1K, improves the base policy from 7.50\% to 27.50\% on RoboCasa, and raises real-world success from 32.5\% to 57.5\% after two policy-update rounds on Astribot S1.

\noindent \textbf{Limitations.}
Despite its effectiveness, MobiAgent has several limitations. First, adding new skills to the shared VLA backbone may interfere with previously learned behaviors. Parameter-efficient modules such as LoRA~\citep{hu2021lora}, together with continual-learning strategies~\citep{kirkpatrick2017overcoming}, may support more stable and non-destructive skill expansion. Second, the Inner Loop still exhibits pose-correction churn, inaccurate placement recovery, and critic drift under transient visual occlusions. Explicit geometric perception, including object pose estimation~\citep{wen2024foundationpose, lin2024sam,liu2025picopose} and monocular geometry~\citep{yang2024depth, liu2026foundationgeo}, could improve pose-sensitive execution, while future--reality verification~\citep{wang2026trust} may enhance critic calibration and failure-triggered replanning. Finally, although MobiAgent is evaluated across RoboCasa, BEHAVIOR-1K, and real-world tasks, broader task families, environments, and robot embodiments are needed to further assess its scalability and generalization.

\newpage
\acknowledgments{
The work has been supported by Hong Kong Research Grant Council - General Research Fund Scheme (Grant No. 17202422, 17212923, 17215025) Theme-based Research (Grant No.T45-701/22-R), and Strategic Topics Grant (Grant No.STG3/E-605/25-N). Part of the described research work is conducted in the JC STEM Lab of Robotics for Soft Materials funded by The Hong Kong Jockey Club Charities Trust.
}

\bibliography{references}

\clearpage
\addtocontents{toc}{\protect\setcounter{tocdepth}{2}}
{\small\setlength{\parskip}{0pt}\hypersetup{linkcolor=black}\tableofcontents}
\clearpage
\setcounter{figure}{0}
\setcounter{table}{0}
\renewcommand{\thefigure}{S\arabic{figure}}
\renewcommand{\thetable}{S\arabic{table}}
\renewcommand{\theHfigure}{supp.\arabic{figure}}
\renewcommand{\theHtable}{supp.\arabic{table}}
\appendix

\section{LLM/VLM Configurations and Prompt Design}
\label{app:prompts}

\Cref{tab:hparams_vlm} details the default foundation models powering each module within the MobiAgent framework. While most components utilize GPT-5.4~\citep{openai2025gpt54}, the Data Curator specifically relies on Gemini-3.1-Pro~\citep{googledeepmind2026gemini31pro} for trajectory segmentation,  driven by the need to accurately localize skill boundaries over continuous video. Because the GPT-5.4 endpoint does not natively support continuous video input, approximating the recording via uniformly sampled frames markedly degrades temporal boundary localization. In contrast, Gemini-3.1-Pro accepts native video input, ensuring precise temporal segmentation.

\begin{table}[h]
  \centering
  \caption{Default LLM/VLM models used in our MobiAgent.}
  \label{tab:hparams_vlm}
  \small
  \setlength{\tabcolsep}{8pt}
  \begin{tabular}{ll}
    \toprule
    Agent & Model \\
    \midrule
    Task Planner                    & GPT-5.4 \\
    Reflection Critic               & GPT-5.4 \\
    Data Curator (segmentation)     & Gemini-3.1-Pro \\
    Data Curator (verification)     & GPT-5.4 \\
    Skill Generator (aggregation)   & GPT-5.4 \\
    Skill Trainer                   & GPT-5.4 \\
    \bottomrule
  \end{tabular}
\end{table}

For the prompt-driven modules, we employ a unified, framework-level prompt design. Crucially, these prompts are strictly task-agnostic: they specify the canonical skill set (i.e., the \texttt{stage\_hint} used for policy routing) and the verb-frame skill templates, but explicitly exclude any specific task, room, or object names from the evaluation suite. Instance-specific context is dynamically injected solely through the goal, history, and observation fields. This zero-shot design ensures that the exact same prompts are reused without modification across all tasks.

In the following subsections, we provide the complete prompt templates for all LLMs and VLMs utilized across both the inner loop (Task Planner and Reflection Critic) and the outer loop (Data Curator for segmentation and verification, and Skill Generator for skill aggregation). Furthermore, all trailing JSON schemas are strictly enforced via constrained decoding to guarantee reliably formatted outputs.

\subsection{Task Planner}
\label{app:prompt_plan}

\begin{promptbox}[System prompt]

You are an online planner for a mobile-manipulation robot. Given the goal,
execution history, current camera observations, and robot state, choose one
next atomic subtask. You are called again after subtask completion or a
failure-triggered replan, so the next step must reflect the current scene
rather than blindly resume an earlier plan.\\[2pt]
INPUTS\\
\hspace*{1em}global\_goal: <task instruction>\\
\hspace*{1em}history: <executed subtasks, outcomes, chunks spent, and Critic evidence>\\
\hspace*{1em}observations: <current scene and gripper-camera images>\\
\hspace*{1em}robot\_state: <available embodiment and proprioceptive information>\\[2pt]
SKILLS AND INSTRUCTION TEMPLATES\\
\hspace*{1em}\texttt{move\_to}: Navigate to the <target>.\\
\hspace*{1em}\texttt{pick\_up\_from}: Pick up the <object> from the <source>.\\
\hspace*{1em}\texttt{place\_in}/\texttt{place\_on}: Place the <object> in/on the <destination>.\\
\hspace*{1em}\texttt{open}/\texttt{close}: Open/Close the <target>.\\
\hspace*{1em}\texttt{open\_close}: Turn on/off the <control>, or Press the <control>.\\
\hspace*{1em}\texttt{action}: <supported process instruction>.\\[2pt]
OUTPUT (JSON)\\
\hspace*{1em}\{ "id": <string>,\\
\hspace*{1em}~~"plan\_sketch": [<remaining atomic instructions>],\\
\hspace*{1em}~~"prompt": <instruction matching plan\_sketch[0]>,\\
\hspace*{1em}~~"stage\_hint": <supported skill>,\\
\hspace*{1em}~~"target\_object\_name": <string|null>,\\
\hspace*{1em}~~"rationale": <one short sentence> \}\\[2pt]
RULES\\
\hspace*{1em}1.~Cover every unfinished goal requirement, including repeated instances. Count only history entries marked complete; an earlier plan sketch is not an execution record. Execute only the first remaining step.\\
\hspace*{1em}2.~Respect the embodiment's reach and grasp capacity. Finish transporting or placing a held object before acquiring another.\\
\hspace*{1em}3.~Navigate only when the next manipulation requires a different station; omit redundant travel and consecutive navigation steps.\\
\hspace*{1em}4.~On a replan, use the current post-failure scene to identify the object's present source and the remaining goal requirements. Do not repeat a completed final step without evidence that its state has regressed.\\
\hspace*{1em}5.~Each instruction expresses one atomic intent. Split acquisition, travel, and placement into separate steps when all are needed; do not combine several skills into one policy prompt.\\
\hspace*{1em}6.~Use concrete object, source, and destination names consistent with the goal and visible scene. Preserve location qualifiers needed to distinguish targets, and keep object identity consistent across related steps.\\
\hspace*{1em}7.~Check prerequisites before selecting a step: the required object must be available, the target reachable, and any necessary access established. Omit already-satisfied prerequisites and do not add actions unrelated to the goal.\\
\hspace*{1em}8.~After a failed attempt, use the Critic's evidence to distinguish an unfinished skill from a wrong target or changed scene. Re-select a supported step that addresses the observed state rather than inventing a recovery skill.\\
\hspace*{1em}9.~The plan sketch contains the remaining sequence, but only its first instruction is executed. Leave retry budgets and overall task termination to the orchestrator. Return JSON only.
\end{promptbox}

\subsection{Reflection Critic}
\label{app:prompt_detect}

\begin{promptbox}[System prompt]

Judge the active subtask from recent labeled camera frames and measured robot
proprioception. Assess the latest state; use earlier frames to establish motion
and progress. Do not assume unobserved scene state.\\[2pt]
INPUTS\\
\hspace*{1em}subtask: \{ prompt, stage\_hint, target\_object\_name \}\\
\hspace*{1em}frames: <labeled scene and gripper views, oldest$\to$newest>\\
\hspace*{1em}robot\_state: <measured gripper aperture and recent base/end-effector motion>\\[2pt]
SUCCESS CONDITIONS\\
\hspace*{1em}Navigation: the robot faces the correct target station within manipulation reach. Seeing the target at a distance or at the edge of a turned-away view is insufficient; corroborate arrival using the local manipulation view.\\
\hspace*{1em}Picking: the named object moves with the gripper, is visibly grasped, or is identified at the fingers together with a stable, nonzero grasp-contact signal. Interpret aperture changes relative to the current attempt; contact with a different object does not satisfy the named subtask.\\
\hspace*{1em}Placing: the object visibly occupies the named destination with support, containment, or a corroborated release sequence. Gripper departure is not required for an established handoff. An object beside the destination, above its boundary without support, or visibly dropped elsewhere is incomplete.\\
\hspace*{1em}Opening/closing: the target panel is visibly in the requested end state, including when it was already in that state.\\
\hspace*{1em}Control actuation: positive evidence of the requested control-state change, such as an indicator, flow, or control-position change. Touching a control or being at its station does not by itself establish completion.\\
\hspace*{1em}Process actions: visible tool--target contact and motion across distinct locations or a direction reversal. A final contact pose or small ambiguous jitter is insufficient evidence of the requested process.\\[2pt]
OUTPUT (JSON)\\
\hspace*{1em}\{ "verdict": "complete" | "incomplete" | "error",\\
\hspace*{1em}~~"reason": <one short sentence>,\\
\hspace*{1em}~~"evidence": [<camera-labeled or measured cues>],\\
\hspace*{1em}~~"recommended\_followup": "next" | "retry" |\\
\hspace*{1em}~~~~"replan\_plan\_deviated" | "replan\_keep\_pose" | "" \}\\[2pt]
RULES\\
\hspace*{1em}1.~Return \texttt{complete} only with positive evidence; default to \texttt{incomplete} under ambiguity. Proximity or gripper motion alone does not establish grasp or placement.\\
\hspace*{1em}2.~Return \texttt{error} for a clearly identified wrong target or persistent physical deadlock. For picking, use \texttt{incomplete} rather than \texttt{error}.\\
\hspace*{1em}3.~Recommend \texttt{next} after completion, \texttt{retry} for unfinished progress, and replanning for a confirmed deviation; leave deadlock recovery to the orchestrator.\\
\hspace*{1em}4.~Cite the camera view or measurement supporting each decisive cue. Use temporal changes to distinguish progress from oscillation or a jam; a single frame does not establish persistent deadlock. Keep the verdict consistent with the evidence.\\
\hspace*{1em}5.~Cross-check available views when the target or contact is occluded. Gripper opening alone does not prove placement, and absence from a view does not prove containment. State uncertainty rather than infer hidden success.\\
\hspace*{1em}6.~Judge only the active subtask's observable completion condition. Do not require it to perform the next skill's transport or retreat phase, and do not treat one completed subtask as proof of overall task success. Return JSON only.
\end{promptbox}

\subsection{Data Curator - Segmenting}
\label{app:prompt_segment}

\begin{promptbox}[System prompt]
Segment a robot-demonstration recording into contiguous atomic-skill clips.
Each clip is exactly one atomic intent: either a TRAVEL (the base navigating
toward a target) or a STATE CHANGE (a manipulation that alters an object's pose
or an articulated part's state); the two are mutually exclusive within a
clip.\\[2pt]
INPUTS\\
\hspace*{1em}recording: <composite head $+$ wrist views; each frame carries a burned-in \texttt{t=...s} playback-time label and the base-velocity readout>\\[2pt]
OUTPUT (JSON)\\
\hspace*{1em}\{ "skill\_timeline": [\\
\hspace*{2em}\{ "segment\_id": <string>,\\
\hspace*{2em}~~"start\_time\_sec": <float>, "end\_time\_sec": <float>,\\
\hspace*{2em}~~"skill\_description": <one-line instruction for this clip> \},\\
\hspace*{2em}\dots ] \}\\[2pt]
RULES\\
\hspace*{1em}1.~Coverage: start at 0.0, consecutive clips touch exactly, end at the last visible \texttt{t=...s}; copy boundaries from the visible time labels.\\
\hspace*{1em}2.~Every gripper acquire$\to$release cycle on a carried object is split into a separate acquisition clip and a separate placement clip, with a navigation clip inserted whenever the base travels between them---never one compound clip.\\
\hspace*{1em}3.~Pre-arrival travel is its own navigation clip; absorb brief approach / align / idle intervals into the adjacent state-change clip.\\
\hspace*{1em}4.~No hallucinated state changes: label a state change only when the change is actually observed, and forbid prep-only labels (\texttt{adjust}, \texttt{reposition}, \texttt{wait}).
\end{promptbox}

\subsection{Data Curator - Verification}
\label{app:prompt_verify}

\begin{promptbox}[System prompt]
You receive a proposed segmentation of one demonstration---an ordered list of
skill segments, each described by a skill description, a stage hint, and a
target object---and decide whether the segmentation is internally consistent
and well-formed for the stated goal. You reason over the textual segmentation
only; you do not watch the recording.\\[2pt]
INPUTS\\
\hspace*{1em}goal: <string>\\
\hspace*{1em}segments: [<ordered list of \{ skill\_description, stage\_hint, target\_object \}>]\\[2pt]
OUTPUT (JSON)\\
\hspace*{1em}\{ "valid": true | false, "reason": <one sentence, $\le$ 25 words> \}\\[2pt]
RULES\\
\hspace*{1em}1.~\texttt{valid}=true only if every segment is a well-formed canonical skill on a plausible target and the ordered sequence is coherent for the goal.\\
\hspace*{1em}2.~\texttt{valid}=false on a malformed segmentation---an incoherent or contradictory ordering, an impossible step, or redundant segments that cannot co-occur.\\
\hspace*{1em}3.~Be conservative---prefer \texttt{valid}=false when the segmentation is ambiguous or under-specified.
\end{promptbox}

\subsection{Skill Generator}
\label{app:prompt_aggregate}

\begin{promptbox}[System prompt]
You receive the free-form skill descriptions produced by segmentation across
many demonstrations. Cluster them by action intent, then give each cluster one
reusable canonical skill name.\\[2pt]
INPUTS\\
\hspace*{1em}items: [\{ item\_id, description \}, \dots] (a flat list, not pre-clustered)\\[2pt]
OUTPUT (JSON)\\
\hspace*{1em}\{ "clusters": [\\
\hspace*{2em}\{ "canonical\_name": <short action-intent label>,\\
\hspace*{2em}~~"activity\_type": "navigation" | "manipulation",\\
\hspace*{2em}~~"item\_ids": [<ids>] \},\\
\hspace*{2em}\dots ] \}\\[2pt]
RULES\\
\hspace*{1em}1.~Cluster by the action performed, not by the task-specific object, room, or support surface.\\
\hspace*{1em}2.~\texttt{canonical\_name} is a short reusable action-intent label.\\
\hspace*{1em}3.~Prefer general categories over fine-grained variants---do not split placing (stack / drop-on / rest-on), acquiring (fetch / retrieve / extract), or actuating (pull-open / swing-shut) into separate names.\\
\hspace*{1em}4.~Fold preparation / contact / completion phases into the state-changing skill they support; never leave them as standalone skills.\\
\hspace*{1em}5.~Every \texttt{item\_id} appears in exactly one cluster.
\end{promptbox}

\section{Implementation Details of Simulated Experiments}

\subsection{Simulation Environment and Embodiment}
\label{app:behavior_setup}

Our simulation experiments are conducted in the OmniGibson simulator using the BEHAVIOR-1K benchmark \cite{li2023behavior1k}. We adopt the Galaxea~R1~Pro, the default embodiment of the 2025 BEHAVIOR Challenge. This wheeled humanoid features an omnidirectional mobile base, a liftable torso, and two 6-DoF arms, making it ideal for tasks requiring extensive end-effector reachability, bimanual coordination, and stable navigation.

The policy receives onboard RGB observations rendered at \(224\times224\) resolution and outputs a \(32\)-dimensional action chunk. Task success is rigorously evaluated using the BDDL goal predicates associated with each activity (\Cref{tab:bddl_verbatim}). For evaluation, every (task, method) pair is averaged across \(10\) random seeds under a fixed scene-reset protocol. Crucially, rather than utilizing long monolithic trajectories, per-skill training clips are extracted from the challenge's teleoperated demonstrations via our Data Curator pipeline.

\subsection{Evaluated Tasks and Subtask Sequences}
\label{app:bddl_instructions}

\begin{table}[h]
  \centering
  \caption{Task instructions and main-paper task names for the four evaluated BEHAVIOR-1K tasks.}
  \label{tab:bddl_verbatim}
  \small
  \setlength{\tabcolsep}{4pt}
  \begin{tabular}{p{0.11\linewidth} p{0.28\linewidth} p{0.53\linewidth}}
    \toprule
    \textbf{Task ID} & \textbf{Task name} & \textbf{Instruction} \\
    \midrule
    \texttt{task-0000} & \texttt{push-radio} &
    ``Turn on the radio on the coffee table.'' \\
    \addlinespace
    \texttt{task-0001} & \texttt{dispose-trash} &
    ``Put the three cans of soda from the living room inside the
    trash can in the kitchen.'' \\
    \addlinespace
    \texttt{task-0016} & \texttt{stack-storage} &
    ``Move the two storage containers from the living room to the
    garage. In the garage, place one container on the floor and stack
    the other container on top of it (either order is fine).'' \\
    \addlinespace
    \texttt{task-0017} & \texttt{fetch-beer} &
    ``Bring two beer bottles from the refrigerator to the coffee
    table, then close the refrigerator.'' \\
    \bottomrule
  \end{tabular}
\end{table}

\Cref{tab:bddl_verbatim} details the natural-language instructions for the four evaluated BEHAVIOR-1K tasks, alongside their abbreviated names used throughout the main text. During an episode, these instruction strings are passed directly into the \textit{global\_goal} field of the Task Planner without modification. Subsequently, \Cref{tab:subtask_trees} outlines the representative per-task subtask sequences used to train the per-skill policies of the Skill Executor in simulation. Each line corresponds to a single segment tagged with one of the canonical skills (\texttt{move\_to}, \texttt{pick\_up}, \texttt{place\_in}, \texttt{place\_on}, \texttt{open}, \texttt{close}). Because the segmentation is highly consistent across episodes, these sequences accurately represent the structural breakdown of each task.

\begin{table}[h]
  \centering
  \caption{Representative verbatim subtask sequences for the four BEHAVIOR-1K tasks.}
  \label{tab:subtask_trees}
  \small
  \setlength{\tabcolsep}{6pt}
  \begin{tabular}{l p{0.74\linewidth}}
    \toprule
    \textbf{Task} & \textbf{Subtask Sequence (verbatim)} \\
    \midrule
    \texttt{push-radio} &
    1.\ move to radio \newline
    2.\ pick up radio from coffee table \newline
    3.\ press radio \newline
    4.\ place radio on coffee table \\
    \addlinespace
    \texttt{dispose-trash} &
    1.\ move to trash can \newline
    2.\ pick up trash can from floor \newline
    3.\ move to can of soda \newline
    4.\ pick up can of soda from floor \newline
    5.\ place can of soda in trash can \newline
    6.\ move to can of soda \newline
    7.\ pick up can of soda from floor \newline
    8.\ place can of soda in trash can \newline
    9.\ move to can of soda \newline
    10.\ pick up can of soda from floor \newline
    11.\ place can of soda in trash can \newline
    12.\ place trash can on floor \\
    \addlinespace
    \texttt{stack-storage} &
    1.\ move to door \newline
    2.\ open door \newline
    3.\ move to storage container \newline
    4.\ pick up storage container from floor \newline
    5.\ move to floor \newline
    6.\ place storage container on floor \newline
    7.\ move to storage container \newline
    8.\ pick up storage container from floor \newline
    9.\ move to floor \newline
    10.\ place storage container on storage container \\
    \addlinespace
    \texttt{fetch-beer} &
    1.\ move to refrigerator \newline
    2.\ open refrigerator \newline
    3.\ pick up beer bottles from refrigerator \newline
    4.\ move to coffee table \newline
    5.\ place beer bottles on coffee table \newline
    6.\ move to refrigerator \newline
    7.\ close refrigerator \\
    \bottomrule
  \end{tabular}
\end{table}

\begin{table}[t]
  \centering
  \caption{Training hyperparameters for the per-skill $\pi_{0.5}$
    fine-tuning in simulation (BEHAVIOR-1K).}
  \label{tab:train_sim_hparams}
  \small
  \setlength{\tabcolsep}{10pt}
  \begin{tabular}{lc}
    \toprule
    \textbf{Setting} & \textbf{Simulation}  \\
    \midrule
    Precision       & bfloat16 \\
    Batch size      & 480  \\
    Training steps  & 50{,}000  \\
    Warmup steps    & 1{,}000  \\
    Learning rate   & $1\times10^{-4}$  \\
    \bottomrule
  \end{tabular}
\end{table}

\subsection{Skill Executor Policy Training}
\label{app:policy_training}

The Skill Executor is instantiated as a single \(\pi_{0.5}\)~\citep{physicalintelligence2025pi0_5} model. This model features a shared PaliGemma (\texttt{gemma\_2b}) backbone and routes to distinct action experts for each canonical skill, where each expert is a \texttt{gemma\_300m} head. All experts are initialized from the publicly available \(\pi_{0.5}\) base checkpoint and subsequently fine-tuned on the skill-specific segment clips generated by the Data Curator pipeline. 

Specifically, the six BEHAVIOR-1K skills (\texttt{move\_to}, \texttt{pick\_up}, \texttt{place\_in}, \texttt{place\_on}, \texttt{open}, \texttt{close}) are fine-tuned jointly on \(8\times\) A100 GPUs using 8-way Fully Sharded Data Parallel (FSDP). We apply a global action normalizer, and the batch shares for each skill are allocated proportionally to their respective dataset sizes. \Cref{tab:train_sim_hparams} summarizes the complete training configurations.

The RoboCasa Skill Executor jointly fine-tunes a shared backbone and six skill-specific action experts using the hyperparameters in \Cref{tab:train_robocasa_hparams}.

\section{Implementation Details of Real-Robot Experiments}

\subsection{Real-Robot Environment and Embodiment}
\label{app:realrobot_setup}

\begin{figure}[t]
  \centering
  \includegraphics[width=\linewidth]{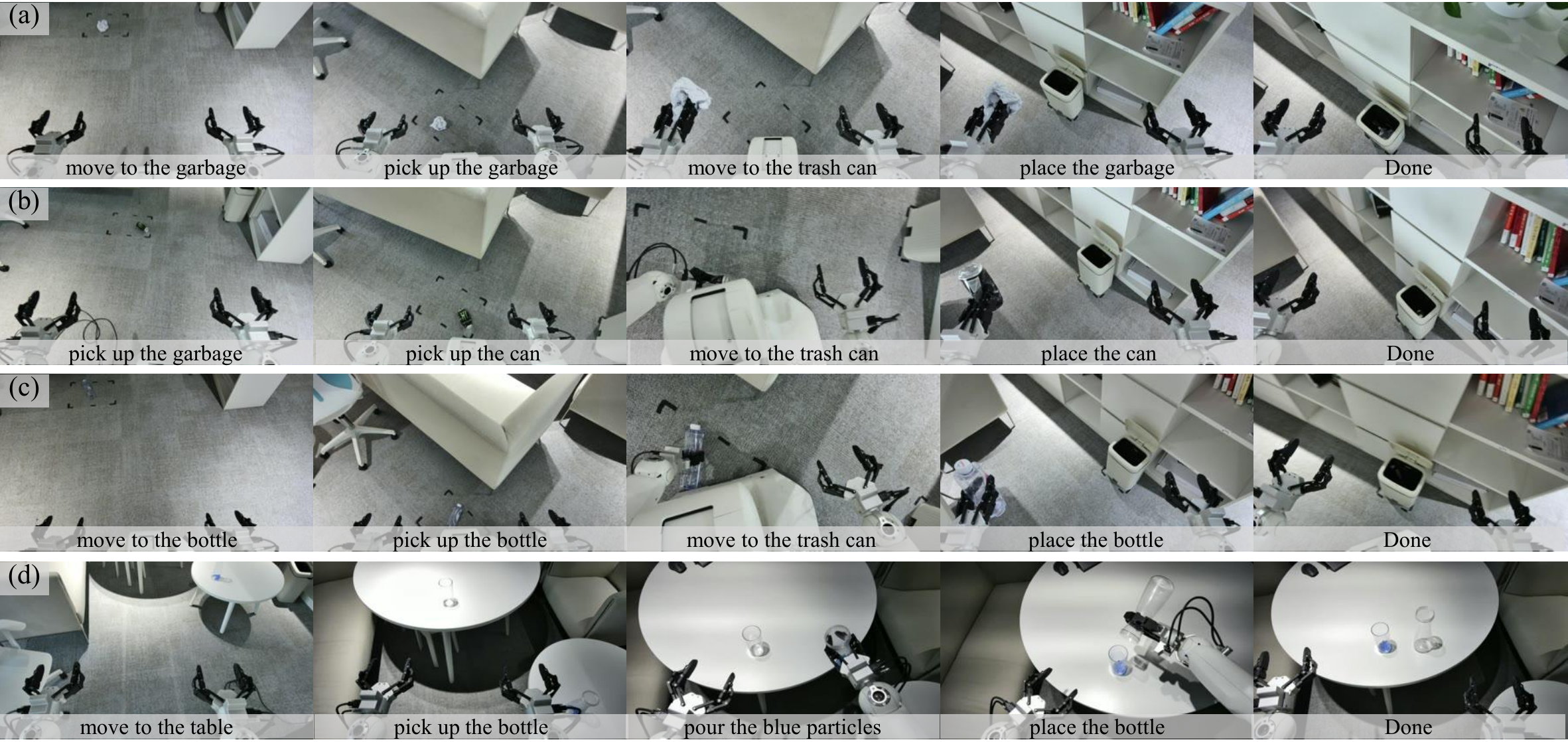}
  \caption{Real-robot rollouts on the Astribot~S1, one task per row, with
    head-camera frames in temporal order (left to right) and the executed
    skill labelled under each frame.
    \textbf{(a)}~\texttt{trash-garbage}: pick up the garbage and place it in
    the trash can.
    \textbf{(b)}~\texttt{trash-can}: pick up the can and place it in the
    trash can.
    \textbf{(c)}~\texttt{trash-bottle}: pick up the bottle and place it in
    the trash can.
    \textbf{(d)}~\texttt{pour-blue}: pick up the bottle and pour the blue
    particles into the cup.}
  \label{fig:real_robot_rollouts}

  \par\vspace{8pt}
  \begingroup
  \makeatletter\def\@captype{table}\makeatother
  \centering
  \caption{Training hyperparameters for the RoboCasa Skill Executor.}
  \label{tab:train_robocasa_hparams}
  \small
  \setlength{\tabcolsep}{8pt}
  \begin{tabular}{ll}
    \toprule
    \textbf{Setting} & \textbf{RoboCasa} \\
    \midrule
    Precision & bfloat16 \\
    Batch size & 48 \\
    Training steps & 25{,}000 \\
    Warmup steps & 1{,}000 \\
    Learning rate & $2.5\times10^{-5}$ \\
    \bottomrule
  \end{tabular}
  \endgroup
\end{figure}

\paragraph{Platform and Cameras.} The real-robot platform is the
Astribot~S1~\citep{gao2025astribot}, a mobile bimanual manipulator
with an omnidirectional wheeled base (3~DoF), a 4-DoF articulated
torso, two 7-DoF arms, and a 2-DoF head (23~DoF in total); each arm
ends in a parallel-jaw gripper rated to a 5\,kg payload. The onboard
RGB-D suite comprises a head camera (stereo RGB plus an Orbbec
Femto~Bolt) as the primary scene view, two wrist-mounted Intel
RealSense~D401 cameras for active-chunk close-ups, and a chest-mounted
Orbbec Gemini~335; all stream at 30\,Hz. Policy actions are issued at
20\,Hz and tracked by the robot's onboard whole-body controller.

\paragraph{Teleoperation Interface.} Demonstrations are collected with
the S1's native VR teleoperation system~\citep{gao2025astribot}: a
Meta~Quest~3S headset with handheld controllers drives the robot
through paired first-person (fine manipulation) and third-person
(whole-body, low-latency) views. The operator commands per-arm
end-effector delta poses in an egocentric frame together with base,
torso, head, and gripper targets---a 34-dimensional action with an
SO(3) rotation parameterisation, streamed at 100\,Hz, which is exactly
the action space the Skill Executor's skill policies predict at deployment.
Demonstrations are processed through the same data curation pipeline used in simulation.

\paragraph{Evaluation.} 10 trials per task per method,
evaluated under a fixed scene-reset protocol; success criteria
mirror BDDL goals. Operator interventions are logged for the HIC
metric; an intervention is required only on hard physical safety
violations. \Cref{fig:real_robot_rollouts} shows a time-ordered rollout of
each real-robot task.

\subsection{Real-Robot Evaluated Tasks and Subtask Sequences}
\label{app:real_subtasks}

\Cref{tab:real_tasks_and_subtasks} details the natural-language instructions for the four evaluated real-robot tasks, alongside their abbreviated names used throughout the text. During an episode, these instruction strings are passed directly into the \textit{global\_goal} field of the Task Planner without modification. Subsequently, the table also outlines the verbatim per-task subtask sequences used to train the per-skill policies of the Skill Executor in the real world. Each line corresponds to a single segment tagged with one of the canonical skills (\texttt{move\_to}, \texttt{pick\_up}, \texttt{place\_in}, \texttt{place\_on}, \texttt{pour}).

\begin{table}[h]
  \centering
  \caption{Real-robot tasks and their corresponding subtask sequences. The prompt column is used identically as the Task Planner's deployment prompt and as the per-skill training prompt. The full subtask sequence for each task is given verbatim.}
  \label{tab:real_tasks_and_subtasks}
  \small
  \setlength{\tabcolsep}{6pt}
  \begin{tabular}{@{} l p{0.32\linewidth} p{0.45\linewidth} @{}}
    \toprule
    \textbf{Task ID} & \textbf{Prompt} & \textbf{Subtask Sequence (verbatim)} \\
    \midrule
    \texttt{trash-garbage} & 
    Pick up the garbage and place it in the trash can. & 
    1.\ move to the garbage \newline
    2.\ pick up the garbage \newline
    3.\ move to the trash can while carrying the garbage \newline
    4.\ place the garbage in the trash can \\
    \addlinespace
    
    \texttt{trash-bottle} & 
    Pick up the bottle and place it in the trash can. & 
    1.\ move to the bottle \newline
    2.\ pick up the bottle \newline
    3.\ move to the trash can while carrying the bottle \newline
    4.\ place the bottle in the trash can \\
    \addlinespace
    
    \texttt{trash-can} & 
    Pick up the can and place it in the trash can. & 
    1.\ move to the can \newline
    2.\ pick up the can \newline
    3.\ move to the trash can while carrying the can \newline
    4.\ place the can in the trash can \\
    \addlinespace
    
    \texttt{pour-blue} & 
    Pick up the bottle containing blue particles and pour the blue particles into the cup. & 
    1.\ move to the table \newline
    2.\ pick up the bottle containing blue particles \newline
    3.\ pour the blue particles into the cup \newline
    4.\ place the bottle on the table \\
    \bottomrule
  \end{tabular}
\end{table}

\subsection{Skill Executor Policy Training}
\label{app:policy_training_real}

The Skill Executor is a single
\(\pi_{0.5}\)~\citep{physicalintelligence2025pi0_5} model that exposes one
action expert per canonical skill behind a shared PaliGemma
(\texttt{gemma\_2b}) backbone; each expert is a \texttt{gemma\_300m}
head. All experts are initialised from the public \(\pi_{0.5}\) base
checkpoint and trained on the skill-only segment clips emitted by the
Data Curator pipeline. \Cref{tab:train_hparams} reports the training
configuration. On the Astribot~S1, the Skill Executor is a
single five-skill model (\texttt{move\_to}, \texttt{pick\_up},
\texttt{place\_in}, \texttt{place\_on}, \texttt{pour}) trained on the combined trash and
pour-blue skill clips. Real-robot runs freeze the vision backbone, use a
per-timestamp action normaliser, and fit on \(4\times\) A100 GPUs.

\begin{table}[t]
  \centering
  \caption{Training hyperparameters for the per-skill \(\pi_{0.5}\)
    fine-tuning on the real robot (Astribot~S1).}
  \label{tab:train_hparams}
  \small
  \setlength{\tabcolsep}{12pt}
  \begin{tabular}{lc}
    \toprule
    \textbf{Setting} & \textbf{Real robot (S1)} \\
    \midrule
    Precision       & bfloat16 \\
    Batch size      & 240 \\
    Training steps  & 20{,}000 \\
    Warmup steps    & 100 \\
    Learning rate   & \(5\times10^{-5}\) \\
    \bottomrule
  \end{tabular}
\end{table}

\subsection{Additional Real-Robot Evaluation on Unitree R1D}
\label{app:unitree_r1d}

\begin{figure}[t]
  \centering
  \includegraphics[width=\linewidth,pagebox=cropbox]
    {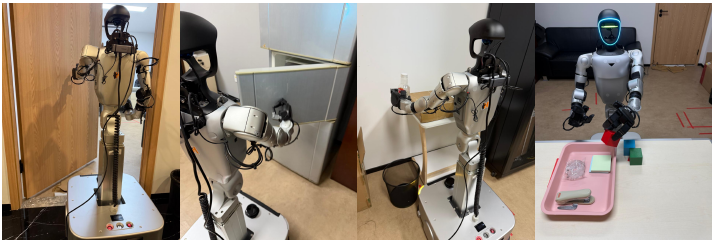}
  \caption{Representative views of the Unitree R1D
    \texttt{home-clean} experiment, illustrating door interaction,
    bottle collection and disposal, and table cleaning.}
  \label{fig:unitree_home_clean}
\end{figure}

We further evaluate MobiAgent on the Unitree R1D
on \texttt{home-clean}, a multi-stage household task involving door
opening, bottle collection and disposal, and table cleaning.
MobiAgent (Round~1) achieves 20.0\% task success, compared with
0.0\% for the base $\pi_{0.5}$ policy.
Representative execution snapshots are shown in
\Cref{fig:unitree_home_clean}.

\section{Additional Quantitative Results and Analysis}
\label{app:additional_results}

This section reports three quantitative analyses that complement the main experiments of \mainsec{4} but were moved out of the main body for space: the wall-clock cost of the Data Curator loop relative to manual annotation, the per-episode runtime cost of the agentic loop relative to the end-to-end baseline, and a detailed per-subtask success rate analysis.

\subsection{Data-collection Wall-clock Cost versus Manual Annotation}
\label{app:data_cost}

\begin{figure}[t]
  \centering
  \includegraphics[width=0.6\linewidth]{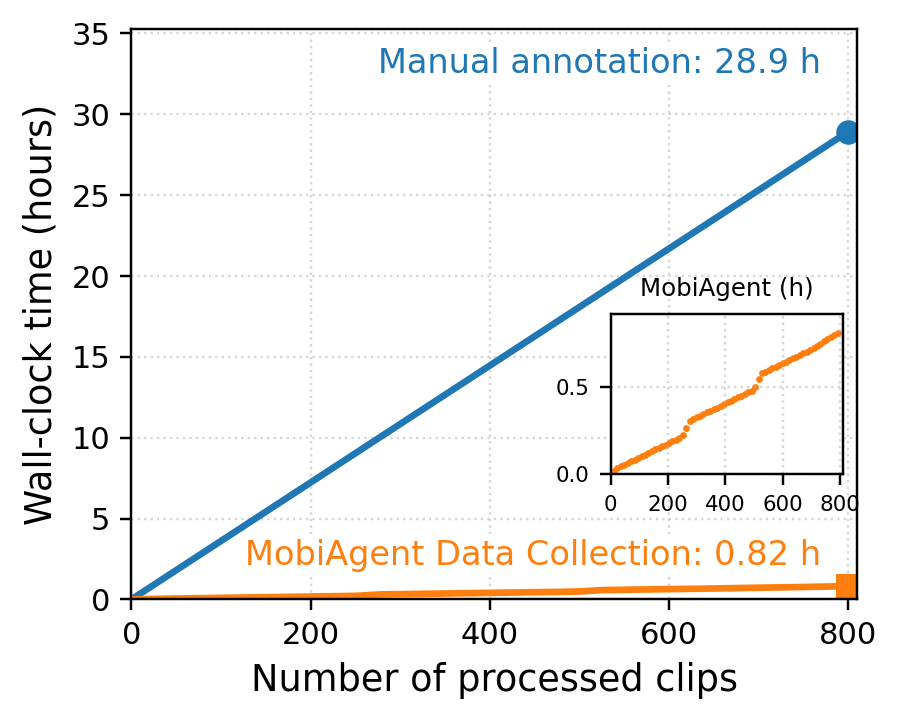}
  \caption{Data-processing wall-clock time vs.\ number of processed clips across all four tasks. At \(800\) clips, manual annotation requires \(\sim 28.9\)\,h compared to MobiAgent's \(\sim 0.82\)\,h (a \(\sim 35{\times}\) speedup).}
  \label{fig:data_efficiency}
\end{figure}

We compare the \emph{data-processing} step that turns a raw demo into a usable per-skill clip (training excluded), starting both sources from the same teleoperated demos at equivalent clip quality. Manual processing is charged at the demo played back at \(2{\times}\) speed (\(\sim 2.2\)\,min per clip); MobiAgent's outer loop is charged at its measured end-to-end API wall-clock on the same demos. At \(800\) clips this is a \(\sim 35{\times}\) reduction (\Cref{fig:data_efficiency}).

\subsection{Per-episode Runtime}
\label{app:runtime}

\Cref{tab:runtime} compares MobiAgent's per-episode runtime against the \(\pi_{0.5}\)-TA end-to-end baseline~\citep{larchenko2025behavior1ksolution} on successful rollouts. For MobiAgent, the total runtime comprises the physical action execution time (policy inference plus physics stepping) and the VLM API time. While MobiAgent's action execution time is moderately higher than the baseline due to running finer-grained skill chunks with per-chunk verification, the primary bottleneck is the VLM API overhead, which is the necessary cost of the closed-loop reasoning that drives its success-rate and robustness gains (\mainsec{4}). To mitigate this latency in future deployments, potential solutions include asynchronous processing, where VLM API calls for verification and high-level planning are executed concurrently with the robot's physical actions, effectively hiding the API latency and significantly reducing the total per-episode runtime.

\begin{table}[h]
  \centering
  \caption{Comparison of per-episode runtime (in seconds) for successful rollouts across the four BEHAVIOR-1K tasks. For MobiAgent, the total runtime is the sum of the action execution time (policy inference and physics stepping, excluding rendering) and the VLM API time. The \(\pi_{0.5}\)-TA baseline operates end-to-end without API calls.}
  \label{tab:runtime}
  \small
  \setlength{\tabcolsep}{10pt}
  \begin{tabular}{@{} l c ccc @{}}
    \toprule
    & \textbf{\(\pi_{0.5}\)-TA} & \multicolumn{3}{c}{\textbf{MobiAgent (ours)}} \\
    \cmidrule(lr){2-2}\cmidrule(l){3-5}
    \textbf{Task} & Total Runtime & Action Exec. & API & Total Runtime \\
    \midrule
    \texttt{push-radio}    &  60 &  81 &  98 &  179 \\
    \texttt{dispose-trash} & 249 & 334 & 381 &  715 \\
    \texttt{stack-storage} & 525 & 707 & 747 & 1454 \\
    \texttt{fetch-beer}    & 538 & 764 & 792 & 1556 \\
    \bottomrule
  \end{tabular}
\end{table}

\subsection{Per-subtask Success Rate}
\label{app:failure_analysis}

\Cref{tab:subtask_sr} reports the Skill Executor's per-subtask success rate by canonical skill, in simulation and on the real robot, helping to localize where the closed loop's residual failures concentrate. 

\begin{table}[t]
  \centering
  \caption{Per-skill success rates (\%) of the Skill Executor in both simulation and real-world deployments, evaluated on instances successfully reached during execution. The ``--'' symbol denotes skills that are exclusive to one domain (e.g., \texttt{open}/\texttt{close} are simulation-only, whereas \texttt{pour} is real-world-only).}
  \label{tab:subtask_sr}
  \small
  \setlength{\tabcolsep}{14pt}
  \begin{tabular}{@{} l cc @{}}
    \toprule
    \textbf{Atomic Skill} & \textbf{Simulation SR (\%)} & \textbf{Real SR (\%)} \\
    \midrule
    \texttt{move\_to}   & 100.00 & 100.00 \\
    \texttt{pick\_up}   &  88.57 &  70.00 \\
    \texttt{place\_in}  &  75.00 &  94.73 \\
    \texttt{place\_on}  & 100.00 & 100.00 \\
    \texttt{open}       &  86.67 &   --   \\
    \texttt{close}      & 100.00 &   --   \\
    \texttt{pour}       &   --   &  50.00 \\
    \bottomrule
  \end{tabular}
\end{table}

\section{Failure Modes and Analysis}
\label{sec:failure_modes}

To better understand the limitations of our system, we categorize the observed execution errors into four distinct failure modes (FM1--FM4), each mapping to specific operational bottlenecks. Specifically, \textbf{FM1 (Reflection Critic False Positives)} involves high-level evaluation errors where the critic incorrectly assumes a subtask is complete (often due to occlusion or timing issues), causing downstream failures. \textbf{FM2 (Bimanual-Coordination Failures)} captures synchronization breakdowns in tightly coupled two-arm tasks, arising because the arms are not explicitly conditioned on each other's states. \textbf{FM3 (Inter-skill Handoff Pose Mismatch)} addresses terminal-pose misalignments during sequential skill transitions, where an off-distribution ending pose of one skill derails the execution of the next. Finally, \textbf{FM4 (Low-level Perceptual Limits)} highlights sensing and representation shortcomings when the system deals with complex spatial layouts or transparent objects.

\subsection{FM1: Reflection Critic False Positives} 

\begin{figure}[h]
  \centering
  \setlength{\tabcolsep}{2pt}
  \begin{tabular}{@{}*{5}{c}@{}}
    \includegraphics[width=0.19\linewidth]{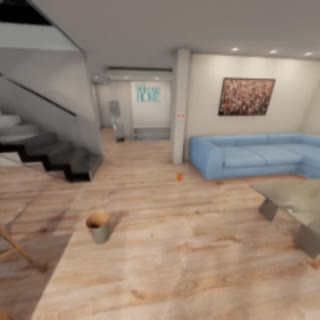} &
    \includegraphics[width=0.19\linewidth]{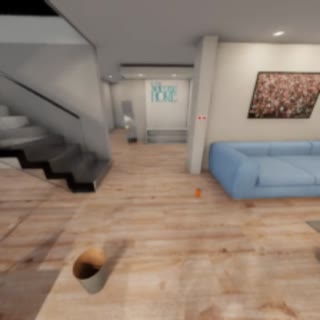} &
    \includegraphics[width=0.19\linewidth]{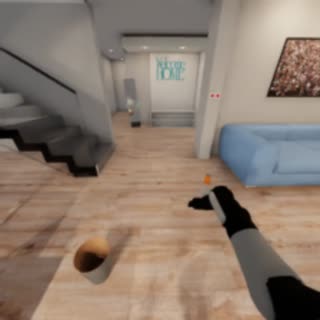} &
    \includegraphics[width=0.19\linewidth]{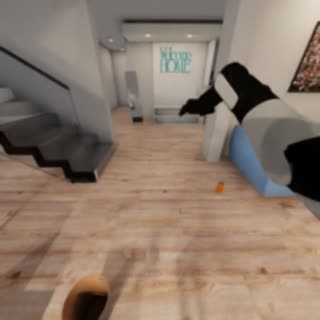} &
    \includegraphics[width=0.19\linewidth]{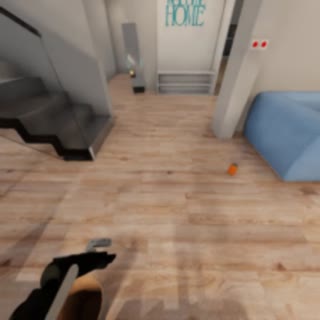} \\
  \end{tabular}
  \caption{FM1: Reflection Critic completion-timing error on \texttt{dispose-trash} (head-camera frames, left to right). Having accepted the \texttt{move\_to} while the robot was still too far, the subsequent \texttt{pick\_up} reaches repeatedly but cannot grasp the out-of-range can, which stays on the floor until the retry budget is exhausted.}
  \label{fig:fail_critic}
\end{figure}

A false positive occurs when the Reflection Critic incorrectly evaluates an incomplete subtask as \texttt{complete}, causing the episode to advance based on a flawed premise. Although the critic's conservative design makes these errors rarer than false negatives, they are significantly more costly because the initial error cascades through all downstream subtasks. A typical manifestation of this issue is a \emph{timing} error, such as terminating a \texttt{move\_to} action prematurely. As illustrated in \Cref{fig:fail_critic}, the critic erroneously approves the robot's approach to the trash can while it is still slightly out of range. Consequently, the subsequent \texttt{pick\_up} action is physically impossible to execute; the robot repeatedly attempts to grasp the can until its retry budget is exhausted, leaving the object untouched on the floor. Fortunately, the system's closed-loop architecture provides a natural fallback: the inevitable failure of the downstream subtask (e.g., the failed \texttt{pick\_up}) will eventually be detected by the Reflection Critic, which will then trigger a re-planning phase to recover from the stalled state.

\subsection{FM2: Bimanual-Coordination Failures}

\begin{figure}[t]
  \centering
  \setlength{\tabcolsep}{2pt}
  \begin{tabular}{@{}*{5}{c}@{}}
    \multicolumn{5}{c}{\small (a) \texttt{fetch-beer}: Poor coordination knocking over the second bottle} \\
    \includegraphics[width=0.19\linewidth]{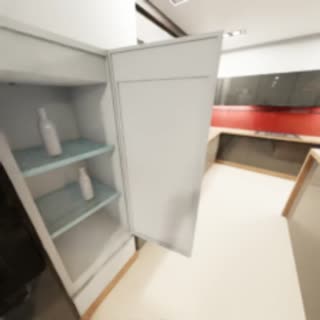} &
    \includegraphics[width=0.19\linewidth]{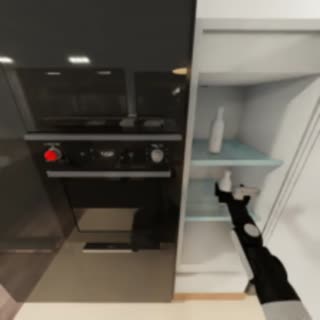} &
    \includegraphics[width=0.19\linewidth]{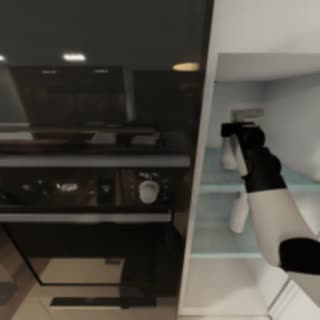} &
    \includegraphics[width=0.19\linewidth]{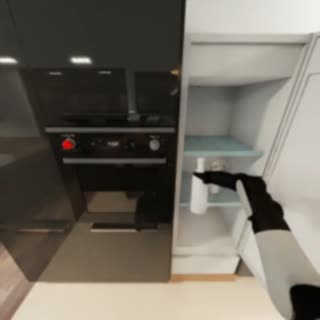} &
    \includegraphics[width=0.19\linewidth]{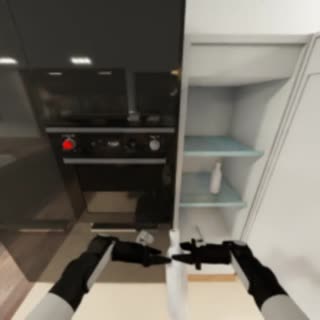} \\
  \end{tabular}
  \vspace{4pt}
  
  \begin{tabular}{@{}*{5}{c}@{}}
    \multicolumn{5}{c}{\small (b) \texttt{push-radio}: Fragile in-hand reorientation for cross-hand press} \\
    \includegraphics[width=0.19\linewidth]{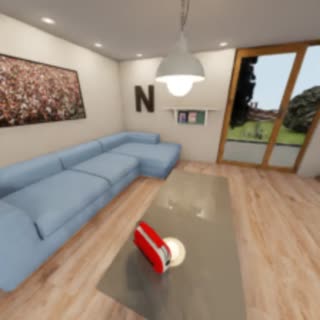} &
    \includegraphics[width=0.19\linewidth]{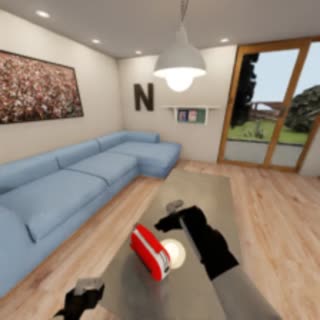} &
    \includegraphics[width=0.19\linewidth]{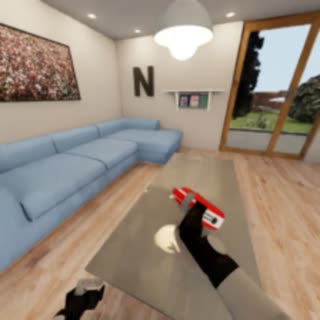} &
    \includegraphics[width=0.19\linewidth]{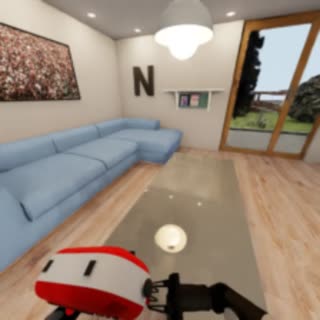} &
    \includegraphics[width=0.19\linewidth]{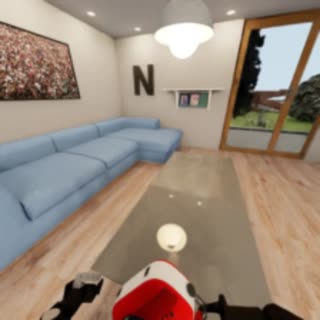} \\
  \end{tabular}
  \vspace{4pt}
  
  \begin{tabular}{@{}*{5}{c}@{}}
    \multicolumn{5}{c}{\small (c) \texttt{stack-storage}: Asymmetric two-arm grasp causing lopsided lift} \\
    \includegraphics[width=0.19\linewidth]{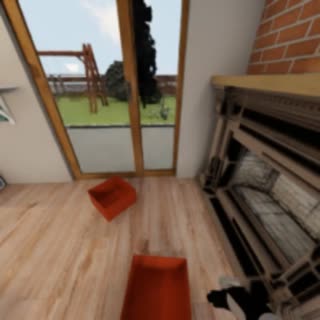} &
    \includegraphics[width=0.19\linewidth]{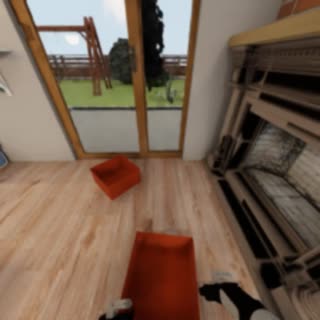} &
    \includegraphics[width=0.19\linewidth]{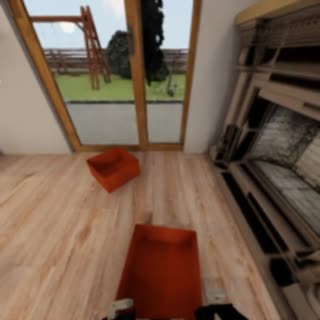} &
    \includegraphics[width=0.19\linewidth]{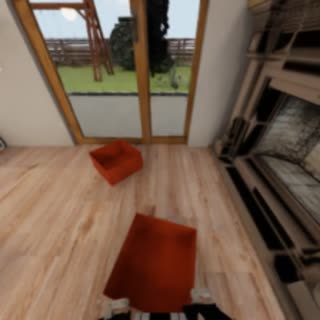} &
    \includegraphics[width=0.19\linewidth]{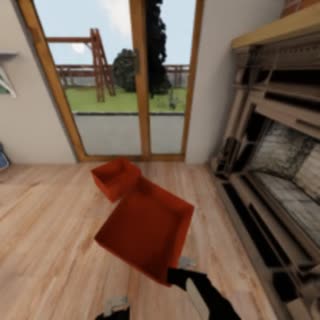} \\
  \end{tabular}
  \caption{FM2: Bimanual-coordination failures across three tasks. (a) The two arms are poorly coordinated, knocking over the target. (b) The picking arm must flip the radio in-hand, frequently dropping it. (c) Only one gripper secures the large container, causing it to tilt and derailing subsequent stacking.}
  \label{fig:fail_bimanual_all}
\end{figure}

Several subtasks require coordinated bimanual actions. However, because the Skill Executor drives both arms from a single head without explicitly conditioning on the partner arm's state, tightly coupled two-arm motions often suffer from poor synchronization. As illustrated in \Cref{fig:fail_bimanual_all}, three recurring failure modes highlight this limitation: 
(a) In \texttt{fetch-beer}, the robot must hold one bottle in one gripper while retrieving a second from the refrigerator. The lack of coordination often causes the reaching arm to knock over the second bottle rather than grasping it cleanly. 
(b) In \texttt{push-radio}, the picking arm grasps the radio in an arbitrary, convenient orientation, which frequently leaves the power button facing away from the pressing arm. This necessitates a fragile in-hand reorientation or hand-over, making it a common source of dropped or misaligned radios. 
(c) In \texttt{stack-storage}, the storage container is too large for a single gripper and requires a simultaneous bimanual grasp. When only one gripper secures it, the container is lifted unevenly, which subsequently derails the carry (\texttt{move\_to}) and stacking (\texttt{place\_on}) subtasks. 
In future work, this limitation could potentially be addressed by introducing more constrained, arm-specific subtask descriptions and generating corresponding targeted training data to better supervise tightly coupled bimanual coordination.

\subsection{FM3: Inter-skill Transition and Pose Mismatch} 

\begin{figure}[t]
  \centering
  \setlength{\tabcolsep}{2pt}
  \begin{tabular}{@{}*{5}{c}@{}}
    \includegraphics[width=0.19\linewidth]{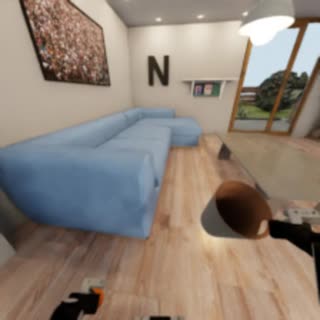} &
    \includegraphics[width=0.19\linewidth]{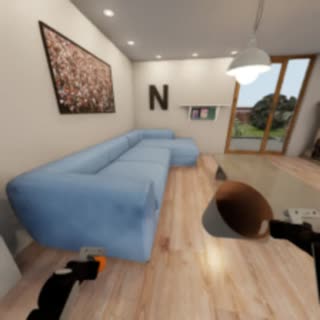} &
    \includegraphics[width=0.19\linewidth]{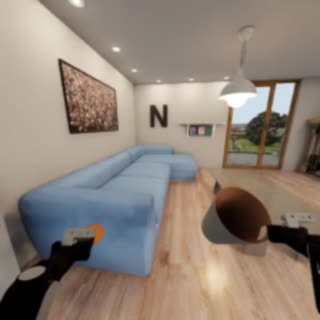} &
    \includegraphics[width=0.19\linewidth]{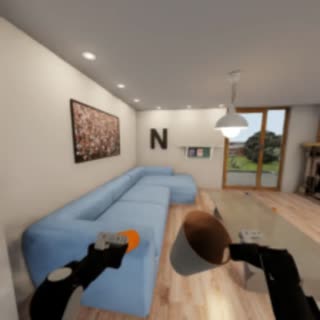} &
    \includegraphics[width=0.19\linewidth]{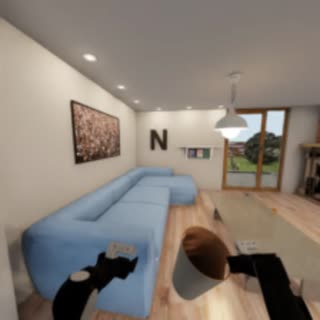} \\
  \end{tabular}
  \caption{FM3: Inter-skill handoff pose mismatch on \texttt{dispose-trash} (head-camera frames, left to right). The \texttt{pick\_up} grasps the can off the floor and succeeds, but ends with the arm raised and rotated away from the carried bin; the next \texttt{place\_in} inherits this pose, cannot bring the can over the bin opening, and the placement fails.}
  \label{fig:fail_handoff}
\end{figure}

A primary challenge in modular policy execution lies in the \emph{skill transition} phase. While training per-skill policies in isolation significantly improves learning efficiency, it inherently assumes a roughly canonical starting configuration for each skill. Consequently, during the transition between skills, if the preceding skill achieves its functional goal but leaves the manipulator in an out-of-distribution (OOD) pose, the subsequent skill inherits this unfamiliar state and may fail despite no explicit error occurring upstream. \Cref{fig:fail_handoff} illustrates this transition bottleneck in a \texttt{dispose-trash} episode: the \texttt{pick\_up} skill successfully grasps the soda can (top row) but terminates in an awkward pose, raised and rotated away from the bin. The subsequent \texttt{place\_in} skill (bottom row) initializes from this misaligned configuration and fails to navigate the can over the bin opening. Because the Reflection Critic currently evaluates each subtask strictly against its isolated goal, the successful \texttt{pick\_up} is accepted, and the kinematic deviation only manifests during the transition to the next step. 

To address this transition bottleneck in future work, the agent framework could be enhanced with transition-aware mechanisms. At the system level, we could dynamically insert a transitional \texttt{reset\_pose} or \texttt{reorient} skill between highly misaligned subtasks to bridge the kinematic gap. Alternatively, the Reflection Critic could be augmented to evaluate not just subtask completion, but also the kinematic viability of the handoff state for the upcoming transition. At the policy level, injecting starting-state noise during skill training would also improve the downstream policy's robustness to imperfect transitions.

\subsection{FM4: Low-level Perceptual Limits} 

\begin{figure}[h]
  \centering
  \setlength{\tabcolsep}{2pt}
  \begin{tabular}{@{}*{4}{c}@{}}
    \multicolumn{4}{c}{\small (a) \texttt{trash-bottle}: Weak spatial perception misjudging bin position} \\
    \includegraphics[width=0.24\linewidth]{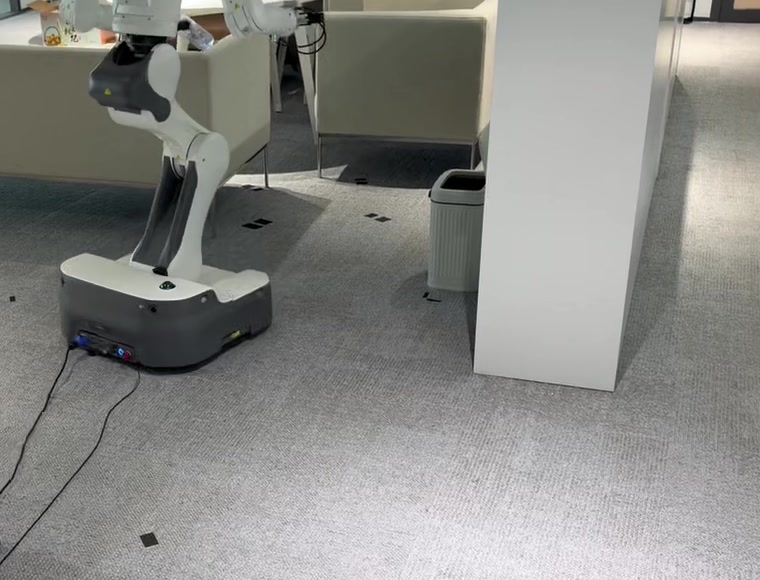} &
    \includegraphics[width=0.24\linewidth]{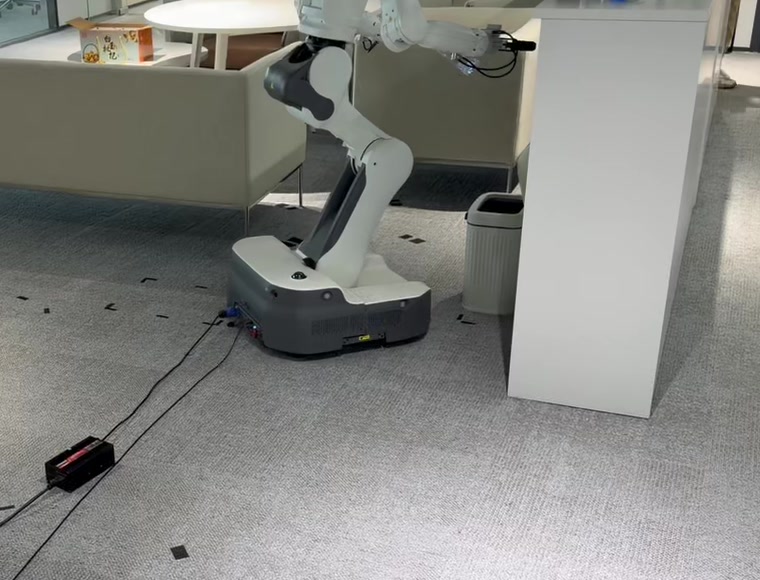} &
    \includegraphics[width=0.24\linewidth]{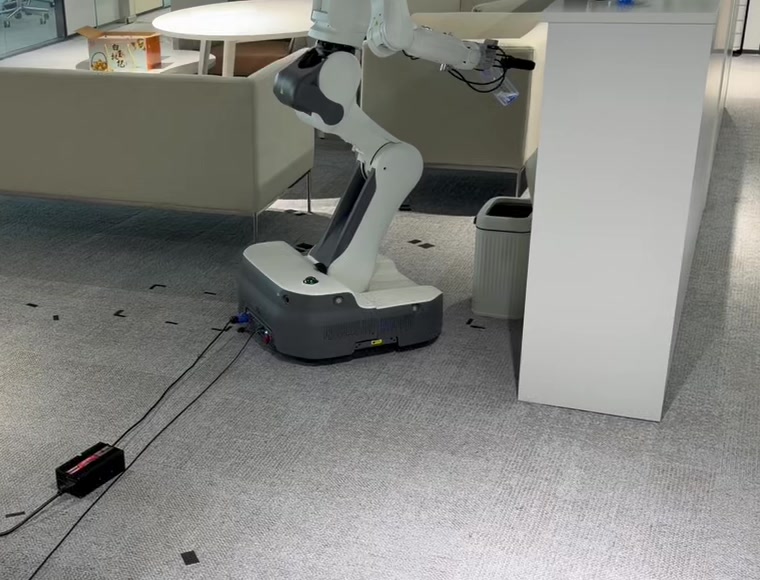} &
    \includegraphics[width=0.24\linewidth]{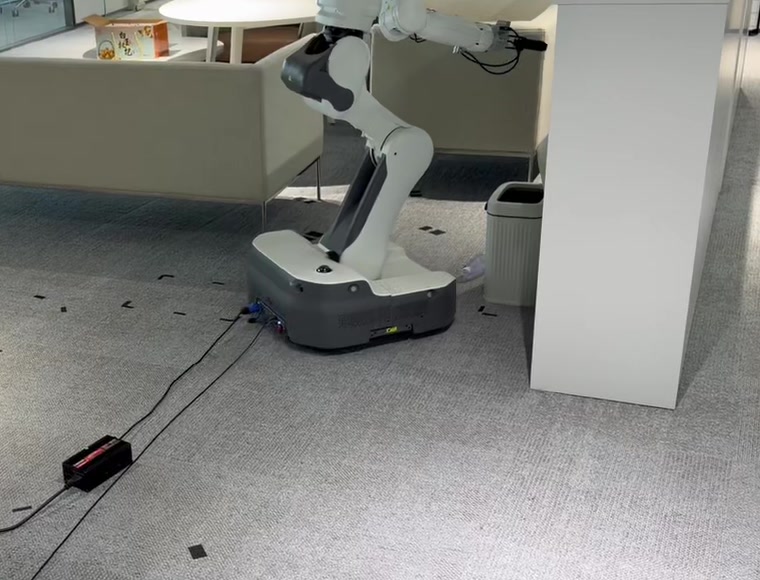} \\
  \end{tabular}
  \vspace{4pt}
  
  \begin{tabular}{@{}*{4}{c}@{}}
    \multicolumn{4}{c}{\small (b) \texttt{pour-blue}: Transparent-object perception failure} \\
    \includegraphics[width=0.24\linewidth]{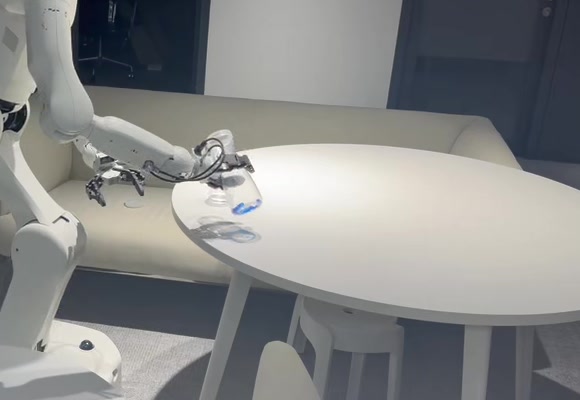} &
    \includegraphics[width=0.24\linewidth]{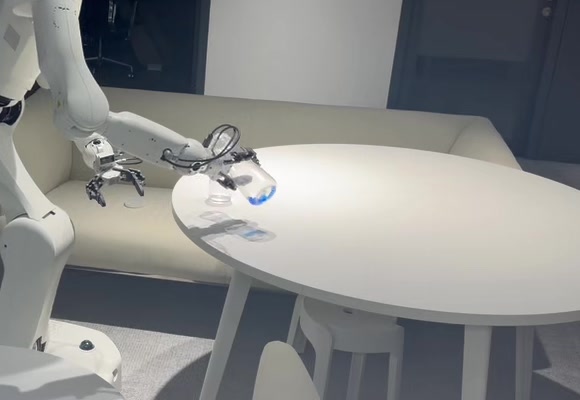} &
    \includegraphics[width=0.24\linewidth]{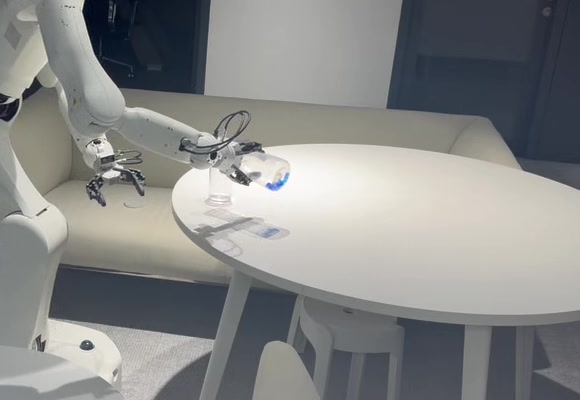} &
    \includegraphics[width=0.24\linewidth]{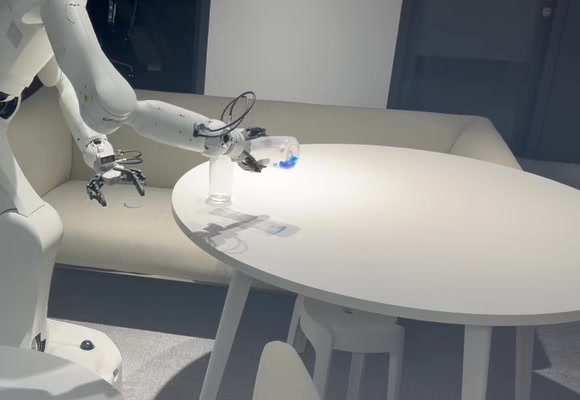} \\
  \end{tabular}
  \caption{FM4: Low-level perceptual limits on the real robot. (a) The policy misjudges the bin's spatial configuration, dropping the bottle on the floor instead of inside. (b) The clear glass is hard to localise, causing the liquid to be poured onto the table beside the glass.}
  \label{fig:fail_perception}
\end{figure}

The shared per-skill policies inherit the perceptual limitations of the underlying action expert, which prominently surface in two real-robot tasks (see \Cref{fig:fail_perception}). In \texttt{trash-bottle} (a), inadequate spatial and metric perception causes the policy to misjudge the relative position of the trash bin, leading the robot to release the bottle just outside the rim so that it falls onto the floor. In \texttt{pour-blue} (b), the clear glass is difficult to localize against the bright table surface; consequently, the policy misaligns the bottle and pours the liquid next to the glass rather than into it. Both failures stem fundamentally from the sensing and representational bottlenecks of the current action expert. Moving forward, these limitations could be addressed by enhancing the dedicated perception module (e.g., incorporating explicit 3D representations or depth sensing). Furthermore, owing to the modular design of our agent framework, any future enhancements in state-of-the-art Vision-Language-Action (VLA) policies can be seamlessly integrated, allowing the system to effortlessly inherit stronger perceptual and spatial reasoning capabilities without requiring architectural overhauls.

\section{Comparison with Other Agentic Frameworks}
\label{app:framework_comparison}

\subsection{Comparison with CaP-X} 

CaP-X~\citep{fu2026capx} is a concurrent
Code-as-Policy framework: a coding agent (a VLM) synthesises an executable
program that composes perception primitives (open-vocabulary segmentation
and pointing) and control primitives to drive the robot. On its mobile
embodiment the only manipulation primitive is a grasp:
\texttt{sample\_grasp\_pose} runs a learned 6-DoF grasp-pose network
(Contact-GraspNet~\citep{sundermeyer2021contactgraspnet}) on the segmented
object's depth point cloud, falling back to top-down / bounding-box
heuristic grasps when the network returns none, and the program tries the
returned candidates one at a time (motion-plan to the pre-grasp pose, close
the gripper, lift) until one holds. Its mobile primitive set is thus just
navigation, this one-shot grasp, and gripper open/close: it exposes no
\texttt{place}, \texttt{open}/\texttt{close}, or \texttt{press} primitive, so
it cannot release into a receptacle, stack an object, or actuate an
articulated object, and therefore cannot carry a long-horizon task to
completion. CaP-X's own
BEHAVIOR evaluation reflects this: it reduces its two mobile tasks to
``pick up the radio'' and ``pick up a soda can'' (navigate $+$ grasp only).
We accordingly compare on exactly the navigate-and-grasp portion of those two
tasks (\texttt{push-radio} and \texttt{dispose-trash}). Following
CaP-X's own protocol we report navigation success (the robot reaches within
$\sim$1\,m of the target) and task (grasp) success separately, over $10$
trials per task under matched observations and embodiment (\Cref{tab:capx}).
Our $10$ trials use \emph{held-out test scenes} unseen in training, whereas
CaP-X's original $25$-scene protocol drew from the training split; since
MobiAgent is learned and CaP-X is training-free, this makes the comparison
conservative for us. CaP-X navigates reliably but grasping is its bottleneck;
MobiAgent both navigates and grasps reliably and additionally executes the
subsequent \texttt{press}/\texttt{place\_in} steps that complete each task,
which CaP-X cannot attempt.

\begin{table}[t]
  \centering
  \caption{Comparison with CaP-X~\citep{fu2026capx} on the two tasks within its
    navigate-and-grasp repertoire ($10$ trials each, held-out test scenes,
    matched observations and embodiment). Navigation and task (including navigation) success
    are reported separately (\%), following CaP-X's protocol.}
  \label{tab:capx}
  \small
  \setlength{\tabcolsep}{10pt}
  \begin{tabular}{l cc cc}
    \toprule
    & \multicolumn{2}{c}{\textbf{CaP-X}~\citep{fu2026capx}}
    & \multicolumn{2}{c}{\textbf{MobiAgent (ours)}} \\
    \cmidrule(lr){2-3}\cmidrule(lr){4-5}
    Task & Navigation SR & Task SR & Navigation SR & Task SR \\
    \midrule
    \texttt{push-radio}    & 100 & 20 & \textbf{100} & \textbf{90} \\
    \texttt{dispose-trash} &  90 & 80 & \textbf{100} & \textbf{100} \\
    \bottomrule
  \end{tabular}
\end{table}

\begin{table}[t]
  \centering
  \caption{RoboClaw's planner decomposition of the four evaluated tasks
    (verbatim subtask prompts from its run logs). Each subtask is a
    coarse composite that already bundles several of MobiAgent's
    atomic skills (e.g.\ \texttt{move\_to}\,$+$\,\texttt{pick\_up}\,$+$%
    \,\texttt{place\_in}).}
  \label{tab:roboclaw_plans}
  \small
  \setlength{\tabcolsep}{4pt}
  \begin{tabular}{l p{0.72\linewidth}}
    \toprule
    \textbf{Task} & \textbf{RoboClaw subtask prompts (verbatim)} \\
    \midrule
    \texttt{push-radio} (2) &
    1.\ Move both arms clear, visually locate the radio on the coffee table in the living room, and position yourself so one gripper can safely reach the radio power control. \newline
    2.\ Use the most suitable gripper to press or toggle the radio's power control on the coffee table until the radio turns on. \\
    \addlinespace
    \texttt{dispose-trash} (3) &
    1.\ Pick up one soda can from the living room, move to the kitchen trash can, and drop the can into the trash. Finish with empty grippers and do not pick a second can in the same run. \newline
    2.\ Pick up one remaining soda can from the living room, move to the kitchen trash can, and drop the can into the trash. Finish with empty grippers and do not pick another can in the same run. \newline
    3.\ Pick up the last remaining soda can from the living room, move to the kitchen trash can, and drop the can into the trash. Finish with empty grippers. \\
    \addlinespace
    \texttt{stack-storage} (2) &
    1.\ Pick up one storage box in the living room, carry or move it to the garage, and place it stably on the garage floor where it can serve as the base for stacking. \newline
    2.\ Pick up the remaining storage box in the living room, move it to the garage, align it above the first box already on the garage floor, and place it stably on top. \\
    \addlinespace
    \texttt{fetch-beer} (5) &
    1.\ Open the fridge door, grasp one bottle of beer from inside the fridge, and take it out securely. \newline
    2.\ Place the beer bottle you are holding onto the coffee table in a stable upright position. \newline
    3.\ Return to the open fridge, grasp a second bottle of beer from inside the fridge, and take it out securely. \newline
    4.\ Place the second beer bottle onto the coffee table in a stable upright position near the first bottle. \newline
    5.\ Close the fridge door fully without disturbing the bottles on the coffee table. \\
    \bottomrule
  \end{tabular}
\end{table}

\subsection{Comparison with RoboClaw} 

We also examine RoboClaw~\citep{roboclaw2026}, a concurrent agentic framework designed for scalable long-horizon tasks. RoboClaw unifies data collection, policy learning, and task execution under a single VLM-driven controller, utilizing Entangled Action Pairs (EAP) to form self-resetting learned policy primitives. Despite its robust multi-policy orchestration, it cannot be applied directly to our skill-conditioned policy due to a fundamental architectural mismatch. RoboClaw's planner decomposes tasks into \emph{coarse}, monolithic subtasks (e.g., an entire ``pick \(\to\) navigate \(\to\) place'' composite issued as a single prompt; see \Cref{tab:roboclaw_plans}). Because a single RoboClaw instruction spans multiple MobiAgent's atomic skills simultaneously, its plans cannot be routed to our skill-conditioned experts without finer re-segmentation. Consequently, we report this architectural divergence rather than a direct head-to-head quantitative comparison.

\paragraph{Static plan-then-execute vs.\ dynamic re-planning.} The two frameworks also differ fundamentally in \emph{when} planning occurs (\Cref{tab:planning_compare}). While RoboClaw's VLM dynamically orchestrates its EAP primitives, at the macro-task level it employs a static, plan-then-execute approach: it generates the full ordered plan upfront and executes it unchanged, only triggering a re-plan if a subtask's success check fails. In contrast, MobiAgent dynamically re-invokes the Task Planner at every step to determine the next subtask based on the current observation. This continuous adaptation, rather than relying solely on failure-triggered corrections, is the architectural driver behind MobiAgent's superior robustness.

\paragraph{Robustness to instruction paraphrasing.} A reliable long-horizon system should produce consistent task decompositions for semantically equivalent instructions. To evaluate this, we hold the task goal fixed and reword its instruction into three distinct paraphrases (\Cref{tab:instr_paraphrases}). We feed each phrasing to both planners using the same GPT-5.4 backbone and starting frame, and record the number and structure of the emitted subtasks. \Cref{tab:instr_plans} reproduces every emitted plan verbatim, while \Cref{tab:instr_robust} summarises the subtask counts.

The results highlight a stark contrast in paraphrase robustness. As quantified in \Cref{tab:instr_robust}, MobiAgent emits a near-constant number of subtasks across different phrasings (with the only minor deviation being a redundant per-bottle cycle elicited by one \texttt{fetch-beer} wording). Its atomic steps remain fixed in number and order; only the object wording adapts to the instruction (e.g., \emph{trash can} vs.\ \emph{kitchen bin}, \emph{fridge} vs.\ \emph{refrigerator}). Conversely, RoboClaw's free-form subtasks fluctuate widely in both count and granularity across all tasks, particularly on \texttt{dispose-trash}.

This discrepancy is rooted in the architectural design. MobiAgent's Task Planner selects each subtask from a fixed catalogue of atomic skills. Thus, the granularity is constrained by the skill set rather than the instruction's surface form---rewording alters the vocabulary \emph{within} a step, but not the overall step count. RoboClaw's free-form generation, however, produces verbose and inconsistent strings that vary with the phrasing. If routed to a policy, these would be far out-of-distribution compared to the short, canonical prompts used during training. By conditioning every expert on a short, fixed skill string (e.g., \texttt{pick up bottle of beer from fridge}), MobiAgent ensures its plans remain paraphrase-robust and its policy prompts strictly in-distribution.

\begin{table}[t]
  \centering
  \caption{Comparison of planning paradigms. RoboClaw employs a static, plan-then-execute approach with conditional recovery, whereas MobiAgent continuously re-plans at every step for dynamic adaptation.}
  \label{tab:planning_compare}
  \small
  \setlength{\tabcolsep}{6pt}
  \begin{tabular}{@{}p{0.22\linewidth} p{0.35\linewidth} p{0.35\linewidth}@{}}
    \toprule
    \textbf{Execution Phase} & \textbf{RoboClaw} & \textbf{MobiAgent (ours)} \\
    \midrule
    \textbf{Task Initialization} &
    Generates the complete, ordered plan upfront. &
    Invokes reactive Task Planner for the immediate subtask (full sketch optional). \\
    \addlinespace
    \textbf{Normal Execution} &
    Executes the fixed plan sequentially without deviation. &
    Re-queries the Task Planner at each step; continuously adapts to observations. \\
    \addlinespace
    \textbf{Error Recovery} &
    Triggers conditional re-planning via an explicit \texttt{replan} action. &
    Inherently recovers by re-evaluating the current state at the next step. \\
    \bottomrule
  \end{tabular}
\end{table}

\begin{table}[t]
  \centering
  \caption{Instruction robustness: Number of subtasks emitted by each planner for the three semantically equivalent phrasings (v0, v1, v2) detailed in \Cref{tab:instr_paraphrases}. The spread \(\Delta = \max - \min\) quantifies consistency across paraphrases. Both planners utilize the same GPT-5.4 backbone and starting frame to isolate the evaluation to the planning decomposition.  paraphrase-robust planner yields a consistent subtask count (i.e., a near-zero \(\Delta\)).}
  \label{tab:instr_robust}
  \small
  \setlength{\tabcolsep}{6pt}
  \begin{tabular}{@{} l cccc cccc @{}}
    \toprule
    & \multicolumn{4}{c}{\textbf{RoboClaw}} 
    & \multicolumn{4}{c}{\textbf{MobiAgent (ours)}} \\
    \cmidrule(lr){2-5}\cmidrule(l){6-9}
    \textbf{Task} & v0 & v1 & v2 & \(\Delta\) & v0 & v1 & v2 & \(\Delta\) \\
    \midrule
    \texttt{push-radio}    &  3 &  2 &  3 & 1 &  4 &  4 &  4 & \textbf{0} \\
    \texttt{dispose-trash} &  6 & 12 &  3 & 9 & 12 & 12 & 12 & \textbf{0} \\
    \texttt{stack-storage} &  4 &  2 &  2 & 2 & 10 &  9 & 10 & \textbf{1} \\
    \texttt{fetch-beer}    &  5 &  4 &  4 & 1 &  7 &  7 & 10 & \textbf{3} \\
    \midrule
    \textbf{Mean \(\Delta\)} & \multicolumn{4}{c}{3.25} 
                             & \multicolumn{4}{c}{\textbf{1.00}} \\
    \bottomrule
  \end{tabular}
\end{table}

\begin{table}[t]
  \centering
  \caption{Instruction paraphrases utilized in the robustness evaluation. For each of the four BEHAVIOR-1K tasks, the underlying goal remains fixed while the instruction is reworded into three semantically equivalent variants (v0, v1, v2). Both planners receive the exact same prompt string for a given variant.}
  \label{tab:instr_paraphrases}
  \small
  \setlength{\tabcolsep}{6pt}
  \begin{tabular}{@{} l l p{0.65\linewidth} @{}}
    \toprule
    \textbf{Task} & \textbf{Variant} & \textbf{Instruction} \\
    \midrule
    \multirow{3}{*}{\texttt{push-radio}}
      & v0 & Turn on the radio sitting on the coffee table in the living room. \\
      & v1 & Switch on the radio that is on the living-room coffee table. \\
      & v2 & Power up the radio resting on the coffee table in the living room. \\
    \addlinespace
    \multirow{3}{*}{\texttt{dispose-trash}}
      & v0 & There are three cans of soda in the living room. Pick up each can and throw it into the trash can in the kitchen, one at a time, until all three cans are in the kitchen trash can. \\
      & v1 & Take the three soda cans from the living room and drop them, one by one, into the kitchen trash can. \\
      & v2 & Clear all three soda cans out of the living room by putting each one into the kitchen bin, handling them one at a time. \\
    \addlinespace
    \multirow{3}{*}{\texttt{stack-storage}}
      & v0 & Move the two storage boxes from the living room to the garage. Place the first box on the garage floor and stack the second box on top of it. \\
      & v1 & Carry both storage boxes from the living room into the garage; set one box down on the garage floor and put the other box on top of it. \\
      & v2 & Relocate the two storage boxes out of the living room into the garage, leaving the first on the floor and stacking the second over it. \\
    \addlinespace
    \multirow{3}{*}{\texttt{fetch-beer}}
      & v0 & Take two bottles of beer from the fridge to the coffee table, then close the fridge door. \\
      & v1 & Bring two beer bottles out of the fridge over to the coffee table, and afterwards shut the fridge door. \\
      & v2 & Move two bottles of beer from the refrigerator onto the coffee table, then close the refrigerator. \\
    \bottomrule
  \end{tabular}
\end{table}

\clearpage
\begingroup
\footnotesize
\setlength{\tabcolsep}{6pt}
\renewcommand{\arraystretch}{1.1}
\begin{longtable}{@{} p{0.09\linewidth} p{0.425\linewidth} p{0.425\linewidth} @{}}
  \caption{Verbatim emitted plans for the instruction-robustness study. This table presents every subtask produced by each planner across the four BEHAVIOR-1K tasks and their three semantically equivalent phrasings (v0, v1, v2) detailed in \Cref{tab:instr_paraphrases}.}
  \label{tab:instr_plans}\\
  \toprule
  \textbf{Variant} & \textbf{RoboClaw} & \textbf{MobiAgent (ours)} \\
  \midrule
  \endfirsthead
  \multicolumn{3}{@{}l}{\small\itshape (Table~\ref{tab:instr_plans}, continued)}\\
  \toprule
  \textbf{Variant} & \textbf{RoboClaw} & \textbf{MobiAgent (ours)} \\
  \midrule
  \endhead
  
  \multicolumn{3}{@{}l}{\textbf{\texttt{push-radio}} (\texttt{task-0000})}\\*[4pt]
  v0 &
      1.\ Move to the living room coffee table and position yourself within reach of the radio while keeping a safe manipulation distance. \newline
      2.\ Visually align an arm and gripper with the radio's power button or power switch so the control can be pressed or toggled cleanly without knocking the radio over. \newline
      3.\ Press or toggle the radio's power control once to turn the radio on, then release contact without moving the radio significantly. &
      1.\ move to radio \newline
      2.\ pick up radio from coffee table \newline
      3.\ press radio \newline
      4.\ place radio on coffee table \\
  \addlinespace
  v1 &
      1.\ Find the radio on the living-room coffee table, identify its power button or switch, and move into a stable pre-press position with a clear approach to the control without activating it yet. \newline
      2.\ Use a careful precise motion to press or toggle the radio's power control on the living-room coffee table and turn the radio on. Stop after the control has been actuated once clearly. &
      1.\ move to radio \newline
      2.\ pick up radio from living-room coffee table \newline
      3.\ press radio \newline
      4.\ place radio on living-room coffee table \\
  \addlinespace
  v2 &
      1.\ Approach the radio on the coffee table in the living room and position your gripper in front of its power control with a stable, collision-free pose ready to actuate it. \newline
      2.\ Actuate the radio's power control on the coffee table to turn the radio on. Press the power button firmly once or rotate the power knob only as much as needed to switch the device from off to on, then release safely. \newline
      3.\ Hold position briefly, observe the radio after actuation, and make any minimal corrective re-press only if needed so the radio is left powered on. &
      1.\ move to radio \newline
      2.\ pick up radio from coffee table \newline
      3.\ press radio \newline
      4.\ place radio on coffee table \\
  \midrule
  
  \multicolumn{3}{@{}l}{\textbf{\texttt{dispose-trash}} (\texttt{task-0001})}\\*[4pt]
  v0 &
      1.\ In the living room, locate one soda can, approach it safely, and grasp it securely with a stable hold for transport. \newline
      2.\ Carry the grasped soda can from the living room to the kitchen trash can, position it over the trash opening, and release it so the can lands inside the trash can. \newline
      3.\ Return attention to the living room, locate a second soda can that is not yet disposed, approach it safely, and grasp it securely with a stable hold for transport. \newline
      4.\ Carry the second grasped soda can from the living room to the kitchen trash can, position it over the trash opening, and release it so the can lands inside the trash can. \newline
      5.\ Locate the last remaining soda can in the living room, approach it safely, and grasp it securely for transport without disturbing the environment unnecessarily. \newline
      6.\ Carry the last grasped soda can to the kitchen trash can, align over the trash opening, and release it so the can lands fully inside the trash can. &
      1.\ move to trash can \newline
      2.\ pick up trash can from floor \newline
      3.\ move to can of soda \newline
      4.\ pick up can of soda from floor \newline
      5.\ place can of soda in trash can \newline
      6.\ move to can of soda \newline
      7.\ pick up can of soda from floor \newline
      8.\ place can of soda in trash can \newline
      9.\ move to can of soda \newline
      10.\ pick up can of soda from floor \newline
      11.\ place can of soda in trash can \newline
      12.\ place trash can on floor \\
  \addlinespace
  v1 &
      1.\ Navigate to the first soda can in the living room and stop in a stable pose that allows grasping. \newline
      2.\ Pick up the first soda can from the living room with a stable grasp and lift it clear of nearby surfaces. \newline
      3.\ Carry the soda can to the kitchen trash can and align it above the trash opening without colliding with the bin or surroundings. \newline
      4.\ Release the soda can so it falls into the kitchen trash can, then retract the gripper without pulling the can back out. \newline
      5.\ Navigate back to the living room and stop in a stable pose near the second remaining soda can for grasping. \newline
      6.\ Pick up the second soda can from the living room with a stable grasp and lift it clear of nearby surfaces. \newline
      7.\ Carry the soda can to the kitchen trash can and align it above the trash opening without colliding with the bin or surroundings. \newline
      8.\ Release the soda can so it falls into the kitchen trash can, then retract the gripper without pulling the can back out. \newline
      9.\ Navigate back to the living room and stop in a stable pose near the third remaining soda can for grasping. \newline
      10.\ Pick up the third soda can from the living room with a stable grasp and lift it clear of nearby surfaces. \newline
      11.\ Carry the soda can to the kitchen trash can and align it above the trash opening without colliding with the bin or surroundings. \newline
      12.\ Release the soda can so it falls into the kitchen trash can, then retract the gripper without pulling the can back out. &
      1.\ move to trash can \newline
      2.\ pick up trash can from floor \newline
      3.\ move to soda can \newline
      4.\ pick up soda can from floor \newline
      5.\ place soda can in trash can \newline
      6.\ move to soda can \newline
      7.\ pick up soda can from floor \newline
      8.\ place soda can in trash can \newline
      9.\ move to soda can \newline
      10.\ pick up soda can from floor \newline
      11.\ place soda can in trash can \newline
      12.\ place trash can on floor \\
  \addlinespace
  v2 &
      1.\ Find one soda can in the living room, pick it up securely, carry it to the kitchen bin, and place it fully inside the bin. \newline
      2.\ Find the next remaining soda can in the living room, pick it up securely, carry it to the kitchen bin, and place it fully inside the bin without disturbing cans already deposited. \newline
      3.\ Find the last remaining soda can in the living room, pick it up securely, carry it to the kitchen bin, and place it fully inside the bin, leaving no soda cans remaining in the living room. &
      1.\ move to kitchen bin \newline
      2.\ pick up kitchen bin from floor \newline
      3.\ move to soda can \newline
      4.\ pick up soda can from floor \newline
      5.\ place soda can in kitchen bin \newline
      6.\ move to soda can \newline
      7.\ pick up soda can from floor \newline
      8.\ place soda can in kitchen bin \newline
      9.\ move to soda can \newline
      10.\ pick up soda can from floor \newline
      11.\ place soda can in kitchen bin \newline
      12.\ place kitchen bin on floor \\
  \midrule
  
  \multicolumn{3}{@{}l}{\textbf{\texttt{stack-storage}} (\texttt{task-0016})}\\*[4pt]
  v0 &
      1.\ Identify one storage box in the living room, reach to it, grasp it securely with both arms if needed, and lift it into a stable carrying pose. \newline
      2.\ Carry the grasped storage box from the living room to the garage and place it flat and stably on the garage floor at a location suitable for stacking a second box on top. \newline
      3.\ Return to the remaining storage box in the living room, grasp it securely with both arms if needed, and lift it into a stable carrying pose. \newline
      4.\ Carry the second storage box from the living room to the garage, align it above the first box already on the garage floor, and place it centered and stably on top of the first box before releasing. &
      1.\ move to door \newline
      2.\ open door \newline
      3.\ move to storage box \newline
      4.\ pick up storage box from floor \newline
      5.\ move to floor \newline
      6.\ place storage box on floor \newline
      7.\ move to storage box \newline
      8.\ pick up storage box from floor \newline
      9.\ move to storage box \newline
      10.\ place storage box on storage box \\
  \addlinespace
  v1 &
      1.\ Pick up one storage box from the living room, carry it into the garage, and place it stably on the garage floor leaving enough free space above it for stacking a second box. \newline
      2.\ Pick up the remaining storage box from the living room, carry it into the garage, align it above the box already on the garage floor, and place it stably on top without tipping or sliding. &
      1.\ move to door \newline
      2.\ open door \newline
      3.\ move to storage box \newline
      4.\ pick up storage box from floor \newline
      5.\ move to storage box \newline
      6.\ pick up storage box from floor \newline
      7.\ move to floor \newline
      8.\ place storage box on floor \newline
      9.\ place storage box on storage box \\
  \addlinespace
  v2 &
      1.\ Pick up one storage box in the living room, carry it into the garage, and place it flat and stably on the garage floor where it can serve as the base for stacking a second box. Release only after the box is stable and not touching obstacles. \newline
      2.\ Pick up the remaining storage box in the living room, carry it into the garage, align it over the first box, and place it stably on top of that box. Release only when the upper box is balanced and fully supported by the lower box. &
      1.\ move to door \newline
      2.\ open door \newline
      3.\ move to storage box \newline
      4.\ pick up storage box from floor \newline
      5.\ move to floor \newline
      6.\ place storage box on floor \newline
      7.\ move to storage box \newline
      8.\ pick up storage box from floor \newline
      9.\ move to storage box \newline
      10.\ place storage box on storage box \\
  \midrule
  
  \multicolumn{3}{@{}l}{\textbf{\texttt{fetch-beer}} (\texttt{task-0017})}\\*[4pt]
  v0 &
      1.\ Open the fridge door, grasp one bottle of beer from inside the fridge, and take it out securely. \newline
      2.\ Place the beer bottle you are holding onto the coffee table in a stable upright position. \newline
      3.\ Return to the open fridge, grasp a second bottle of beer from inside the fridge, and take it out securely. \newline
      4.\ Place the second beer bottle onto the coffee table in a stable upright position near the first bottle. \newline
      5.\ Close the fridge door fully without disturbing the bottles on the coffee table. &
      1.\ move to fridge \newline
      2.\ open fridge \newline
      3.\ pick up bottle of beer from fridge \newline
      4.\ move to coffee table \newline
      5.\ place bottle of beer on coffee table \newline
      6.\ move to fridge \newline
      7.\ close fridge \\
  \addlinespace
  v1 &
      1.\ Open the fridge door wide enough to access the beer bottles inside. \newline
      2.\ Take one beer bottle out of the open fridge and place it upright on the coffee table. \newline
      3.\ Take a second beer bottle out of the open fridge and place it upright on the coffee table next to the first one. \newline
      4.\ Close the fridge door fully. &
      1.\ move to fridge \newline
      2.\ open fridge \newline
      3.\ pick up beer bottle from fridge \newline
      4.\ move to coffee table \newline
      5.\ place beer bottle on coffee table \newline
      6.\ move to fridge \newline
      7.\ close fridge \\
  \addlinespace
  v2 &
      1.\ Open the refrigerator door enough to access the beer bottles inside. \newline
      2.\ Pick up one bottle of beer from inside the refrigerator, carry it carefully, and place it upright on the coffee table. \newline
      3.\ Pick up a second bottle of beer from inside the refrigerator, carry it carefully, and place it upright on the coffee table next to the first bottle. \newline
      4.\ Close the refrigerator door fully after removing the two beer bottles. &
      1.\ move to refrigerator \newline
      2.\ open refrigerator \newline
      3.\ pick up bottle of beer from refrigerator \newline
      4.\ move to coffee table \newline
      5.\ place bottle of beer on coffee table \newline
      6.\ pick up bottle of beer from refrigerator \newline
      7.\ move to coffee table \newline
      8.\ place bottle of beer on coffee table \newline
      9.\ move to refrigerator \newline
      10.\ close refrigerator \\
  \bottomrule
\end{longtable}
\endgroup

\end{document}